\documentclass[lettersize,journal]{IEEEtran}
\usepackage{amsmath,amsfonts}
\usepackage{algorithmic}
\usepackage{algorithm}
\usepackage{array}
\usepackage[caption=false,font=normalsize,labelfont=sf,textfont=sf]{subfig}
\usepackage{textcomp}
\usepackage{stfloats}
\usepackage{url}
\usepackage[colorlinks=true, linkcolor=blue, citecolor=blue, urlcolor=blue]{hyperref}
\usepackage{verbatim}
\usepackage{graphicx}
\usepackage{cite}
\usepackage{booktabs}
\usepackage{multirow}
\usepackage{pifont}
\newcommand{\cmark}{\ding{51}} % √
\newcommand{\xmark}{\ding{55}} % ×
\usepackage{colortbl}     
\usepackage{xcolor}  
\usepackage{amssymb}
\definecolor{lightblue}{RGB}{230,245,255}
\begin{document}

\title{Selectivity Drives Efficiency: Dataset Pruning \\ for Visual Place Recognition}

\author{Tong Jin, Yunpeng Liu, Shuyu Hu, Chun Yuan,~\IEEEmembership{Senior Member,~IEEE,} \\ Song Wang,~\IEEEmembership{Senior Member,~IEEE,} Feng Lu
\thanks{Tong Jin, Yunpeng Liu and Shuyu Hu are with the Shenyang Institute of Automation, Chinese Academy of Sciences, Shenyang 110016, China, and also with University of Chinese Academy of Sciences, Beijing 100049, China (e-mail: jintong@sia.cn; ypliu@sia.cn; hushuyu@sia.cn).}% <-this % stops a space
\thanks{Chun Yuan is with Tsinghua Shenzhen International Graduate School, Tsinghua University, Shenzhen 518071, China (e-mail: yuanc@sz.tsinghua.edu.cn).}
\thanks{Song Wang and Feng Lu are with the Shenzhen University of Advanced Technology, Shenzhen 518107, China (e-mail: lufeng@suat-sz.edu.cn; wangsong@suat-sz.edu.cn).}
\thanks{Feng Lu and Yunpeng Liu are the Corresponding Authors.}
}% <-this % stops a space

% The paper headers
\markboth{Journal of \LaTeX\ Class Files,~Vol.~14, No.~8, August~2021}%
{Shell \MakeLowercase{\textit{et al.}}: A Sample Article Using IEEEtran.cls for IEEE Journals}

\IEEEpubid{0000--0000/00\$00.00~\copyright~2021 IEEE}
% Remember, if you use this you must call \IEEEpubidadjcol in the second
% column for its text to clear the IEEEpubid mark.

\maketitle

\begin{abstract}
Recent visual place recognition (VPR) studies have increasingly relied on large-scale datasets to train more robust and discriminative models. Although this trend significantly improves recognition performance, it also introduces substantial storage and training costs, especially when new architectures or training strategies need to be repeatedly developed and evaluated. Dataset pruning (DP) provides a promising way to improve data efficiency by retaining only informative training data. However, conventional DP methods mainly follow the sample-wise classification paradigm, which overlooks the relation-dependent training nature of VPR, where supervision is typically formed by image pairs rather than independent images. To address this issue, we propose a place-wise dataset pruning framework tailored for VPR. Instead of pruning individual images, our method treats each place as the basic pruning unit and introduces two complementary novel metrics, i.e., intra-place diversity (IPD) and inter-place similarity (IPS), to evaluate the training value of each place. By jointly considering these two metrics, our method ranks all places and constructs a compact yet informative coreset, thereby allowing the pruned dataset to still support the training of robust and discriminative VPR models. Extensive experiments demonstrate that our method consistently outperforms state-of-the-art DP baselines under different pruning ratios while reducing selection and training costs. Moreover, by pruning a merged dataset roughly 3.5$\times$ the size of GSV-Cities to a comparable scale, our coreset maintains highly competitive performance, achieving 94.5\% R@1 on MSLS-val and 97.0\% R@1 on Nordland with only NetVLAD. Codes will be made publicly available.
\end{abstract}

\begin{IEEEkeywords}
Visual place recognition, dataset pruning, coreset selection, efficient training.
\end{IEEEkeywords}

\section{Introduction}
\label{sec:introduction}
\IEEEPARstart{V}{isual} place recognition (VPR) \cite{rvpr,vpr_rs,vprsurvey1,vprbenchmark2} is a fundamental vision task that aims to estimate the approximate location of a query by retrieving the most similar reference from a large geo-tagged database. It serves an essential role in various applications, e.g., robot localization \cite{robotlocalization}, augmented reality \cite{augreality}, and autonomous driving \cite{autodriving}. Due to viewpoint changes, condition variations, and perceptual aliasing \cite{vprsurvey1}, VPR remains challenging. In response, recent studies \cite{gsv_cities,salad-cm,cosplace,selavpr++,megaloc} have used large-scale datasets to train powerful models, substantially mitigating these issues. While effective, the ever-growing volume of data also imposes significant resource requirements, encompassing both training burden and storage overhead.

\begin{figure}[t]
    \centering
    \includegraphics[width=1.0\linewidth]{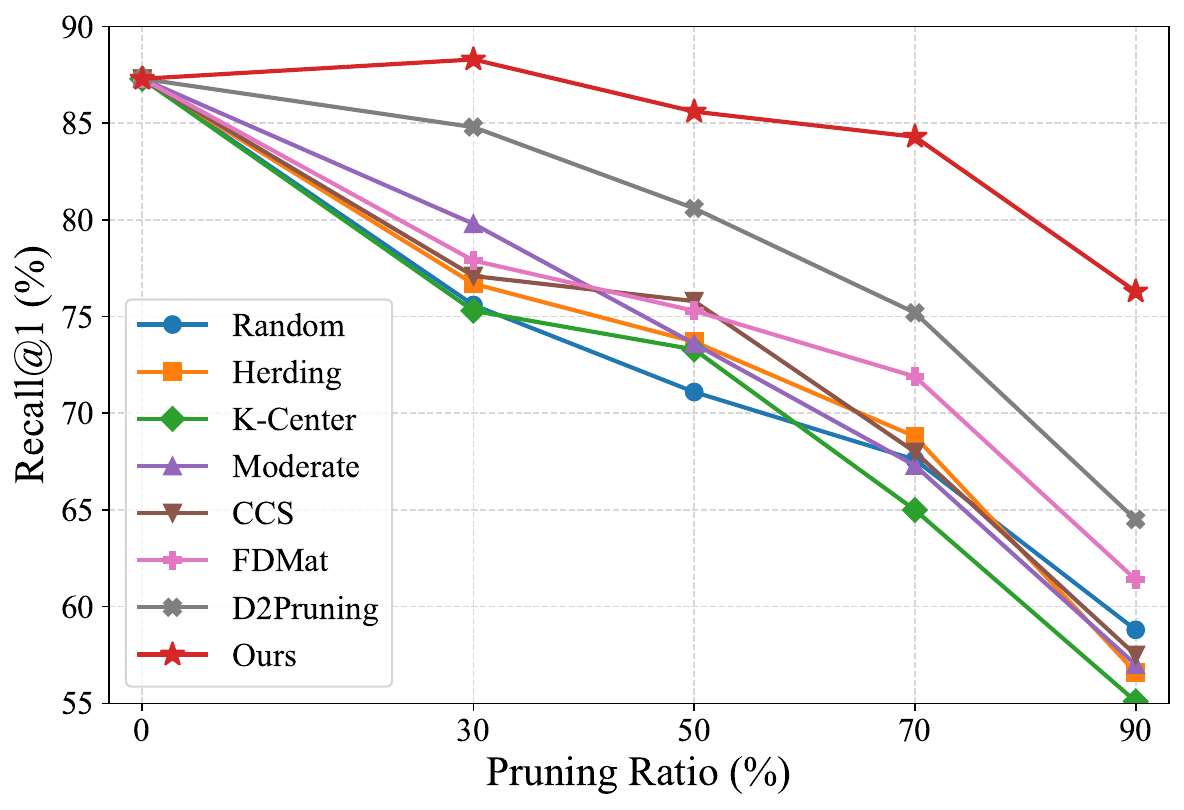}
    \vspace{-0.6cm}
    \caption{Comparison of different dataset pruning methods on Nordland \cite{nordland}. Each pruning method is applied to GSV-Cities \cite{gsv_cities} to select a coreset, on which a VPR model (consistently consisting of the DINOv2-B backbone \cite{dinov2} and the NetVLAD aggregator \cite{netvlad}) is trained. Our method achieves significantly higher Recall@1 than other methods across a range of pruning ratios.}
    \label{fig:performance_comparison}
    \vspace{-0.5cm}
\end{figure}

\IEEEpubidadjcol

Dataset pruning (DP) \cite{forgetting,el2n,infobatch,dyn-unc}, also known as coreset selection \cite{k-center, cdsdfcs, csbocls}, is a practical tool to alleviate such costs. It aims to reduce the size of a large-scale dataset by retaining informative samples, thus accelerating training and improving data efficiency without significantly sacrificing model performance. Existing DP methods \cite{dpsurvey} are mainly developed for classification and can be broadly grouped into several categories. Geometry-based approaches \cite{herding,k-center,moderate_ds,fdmat} promote diversity by removing nearby or redundant samples in the feature space. Uncertainty- and loss/error-based methods \cite{dyn-unc,forgetting,el2n} design score functions to estimate sample difficulty during training. Moreover, several studies \cite{ccs,d2pruning,infomax} attempt to integrate data difficulty and diversity for better selection. These methods have shown promising results in classification-oriented scenarios, where each image is typically treated as an independent training sample. However, this sample-wise paradigm is less aligned with VPR, where the training supervision is usually constructed from pairwise relationships among images rather than from isolated image labels.

Unlike image classification, where each sample is treated as an independent instance contributing individually to model training, VPR is generally a retrieval-based task \cite{vprbenchmark2} that relies on the relational structure between images. Images depicting the same place form positive pairs, while those from different places constitute negative pairs for metric learning. That is, the importance of a single image cannot be meaningfully evaluated in isolation, as its contribution to training depends on how it interacts with other samples to form informative pairs. This dependence on inter-image context results in a significant gap between existing DP methods and their applicability to VPR. Specifically, score-based methods that rely on per-sample training dynamics, such as forgetting events \cite{forgetting} and loss values \cite{el2n}, are not naturally defined at the individual-image level under pairwise VPR supervision. Geometry-based approaches that promote feature diversity may also inadvertently remove multiple images from the same place or visually similar images across places \cite{k-center,d2pruning}, thereby reducing the availability of positive pairs and hard negative pairs. These issues highlight the need for a VPR-oriented DP method.

In this paper, we propose a place-wise dataset pruning framework tailored for VPR. Instead of selecting individual images, our method treats each place as the basic pruning unit and retains/discards places as a whole, which better matches the relational supervision in VPR training. Based on this formulation, we design two complementary metrics to estimate the training value of each place. The first metric, \textit{intra-place diversity} (IPD), measures the visual variation among images within a place and reflects its potential to form informative positive pairs. The second metric, \textit{inter-place similarity} (IPS), measures the visual similarity between a place and its neighboring places, reflecting its potential to provide hard negative pairs. By fusing IPD and IPS into a unified score, our method ranks all places and retains high-score places to construct a compact yet informative coreset. Specifically, on GSV-Cities \cite{gsv_cities}, our method can remove 50\% of training places with negligible performance degradation, with R@1 drops of only 0.1 and 0.5 percentage points on Pitts30k and MSLS-val, respectively. More importantly, when applied to a merged dataset \cite{selavpr++} roughly 3.5$\times$ the size of GSV-Cities, our method selects a coreset with a scale comparable to GSV-Cities while maintaining highly competitive performance against models trained on much larger multi-dataset training pipelines.

Our main \textbf{contributions} can be summarized as follows:

\noindent \textbf{1)} We revisit dataset pruning for VPR from a relation-aware perspective and formulate a place-wise pruning paradigm. This formulation shifts the pruning unit from individual images to places, making dataset pruning better aligned with the pairwise supervision commonly used in VPR training.

\noindent \textbf{2)} We propose a simple and scalable place selection strategy based on two complementary metrics, namely \textit{intra-place diversity} and \textit{inter-place similarity}, which evaluate the training value of each place from intra-place and inter-place perspectives to construct compact and informative coresets.

\noindent \textbf{3)} We demonstrate the scalability and utility of the proposed framework for large-scale VPR training. On the larger merged dataset, our coreset reduces the training data size by 71\% and the total training time by 79.6\%, while providing a compact and reusable subset for developing more VPR models.

\noindent \textbf{4)} Extensive experiments on multiple benchmark datasets demonstrate that our method outperforms SOTA pruning baselines with less selection time. The results on Nordland that highlight the advantages of our method are shown in Fig. \ref{fig:performance_comparison}.

\section{Related Work}
\label{sec:related work}
\subsection{Visual Place Recognition}
VPR is usually formulated as an image retrieval problem, where each image is encoded into a global descriptor and nearest-neighbor search retrieves reference images depicting the same place as the query. Early learning-based VPR methods \cite{netvlad,gem,patch_netvlad} mainly relied on convolutional neural networks and aggregation modules, such as NetVLAD \cite{netvlad} and GeM \cite{gem}, to construct robust global representations. With the development of vision transformers \cite{vit} and large-scale pre-trained models \cite{dinov2}, recent methods improve VPR performance by designing stronger backbones \cite{transvpr,r2former,selavpr,emvp,effovpr,fol,unipr-3d} and more effective aggregation modules \cite{mixvpr,salad,boq,cricavpr,supervlad,qdfl,edtformer,vprcloak}. In addition to model architecture, training data and supervision strategies play a crucial role in learning discriminative place representations. Early works commonly adopted weakly supervised training on Pitts30k \cite{pitts} and MSLS \cite{msls}, where positive and negative pairs are mined according to geographic information. CosPlace \cite{cosplace} and EigenPlaces \cite{eigenplaces} later explored classification-based training on geographically partitioned datasets such as SF-XL \cite{cosplace}. In recent years, GSV-Cities \cite{gsv_cities} has become a widely used fully supervised training dataset for modern VPR models. More recently, large-scale multi-dataset training has been explored to improve generalization, as exemplified by SALAD-CM \cite{salad-cm}, SelaVPR++ \cite{selavpr++}, and MegaLoc \cite{megaloc}. While these training pipelines bring strong recognition performance, they also increase storage and computational costs, motivating more data-efficient VPR training.

\subsection{Dataset Pruning}
With the ever-increasing scale of training data, dataset pruning (DP) has attracted growing attention as a promising way to improve training efficiency. Different from dataset distillation or condensation methods \cite{ddsurvey1} that synthesize new samples, DP preserves real samples from the original dataset, making the selected subset reusable and interpretable. Existing DP approaches can be broadly grouped into three categories. Score-based methods \cite{forgetting,el2n,grad_match,aum,svp,dyn-unc,infobatch} estimate the importance or difficulty of individual images using well-designed scoring criteria. For instance, \textit{Forgetting} \cite{forgetting} counts how often a model’s prediction flips from correct to incorrect during training. \textit{EL2N} \cite{el2n} measures data difficulty by averaging the L2-norm errors across multiple networks, while \textit{Dyn-Unc} \cite{dyn-unc} incorporates training dynamics to refine uncertainty-based scores. Geometry-based methods \cite{herding,k-center,cdal,moderate_ds,cdsdfcs,cal,mcaoal,fdmat} aim to construct a compact subset that maximizes feature diversity. \textit{Herding} \cite{herding} minimizes the discrepancy between the coreset mean and the full dataset mean in feature space. \textit{K-Center} \cite{k-center} ensures wide data coverage, while \textit{Moderate} \cite{moderate_ds} selects samples near the median according to their distances to class centers. Hybrid methods \cite{ccs,d2pruning,infomax} combine score- and geometry-based strategies to achieve a better trade-off. For example, \textit{D2Pruning} \cite{d2pruning} represents the dataset as an undirected graph and performs message passing to consider both difficulty and diversity. Despite their effectiveness in classification tasks, these methods are primarily designed under a sample-wise selection paradigm, making them less aligned with the pairwise supervision in VPR. \textit{In contrast, we perform pruning at the place level and design relation-aware metrics to select informative places for VPR training.}

\subsection{VPR Mining and Data Organization}
Mining strategies and data organization have been widely studied to improve VPR training. These methods mainly refine the labels, pair relationships, or training batches constructed from existing data, rather than directly reducing the size of the training set. For example, EigenPlaces \cite{eigenplaces} reorganizes training images according to viewpoint-related geographic structures, exposing the model to diverse views of the same scene and improving viewpoint robustness. Graded similarity supervision \cite{gcl,gcl2} observes that binary labels are noisy proxies for visual similarity, and relabels image pairs with continuous similarity scores derived from camera pose and overlap metadata, thereby reducing the reliance on costly hard-pair mining and accelerating convergence. CliqueMining \cite{salad-cm} offline reorganizes training data into more challenging batches based on visual similarity to enhance geographic distance sensitivity. Although these methods are closely related to how training data are used in VPR, they address a different problem from dataset pruning. Specifically, they improve the supervision, organization, or batching of the existing dataset, while the overall training set size is generally kept unchanged. \textit{In contrast, our work focuses on reducing the training set itself before model training by selecting a compact subset of informative places. Therefore, our method directly targets data-volume reduction, storage saving, and training-cost reduction.}

\begin{figure*}[t]
    \centering
    \includegraphics[width=1.0\linewidth]{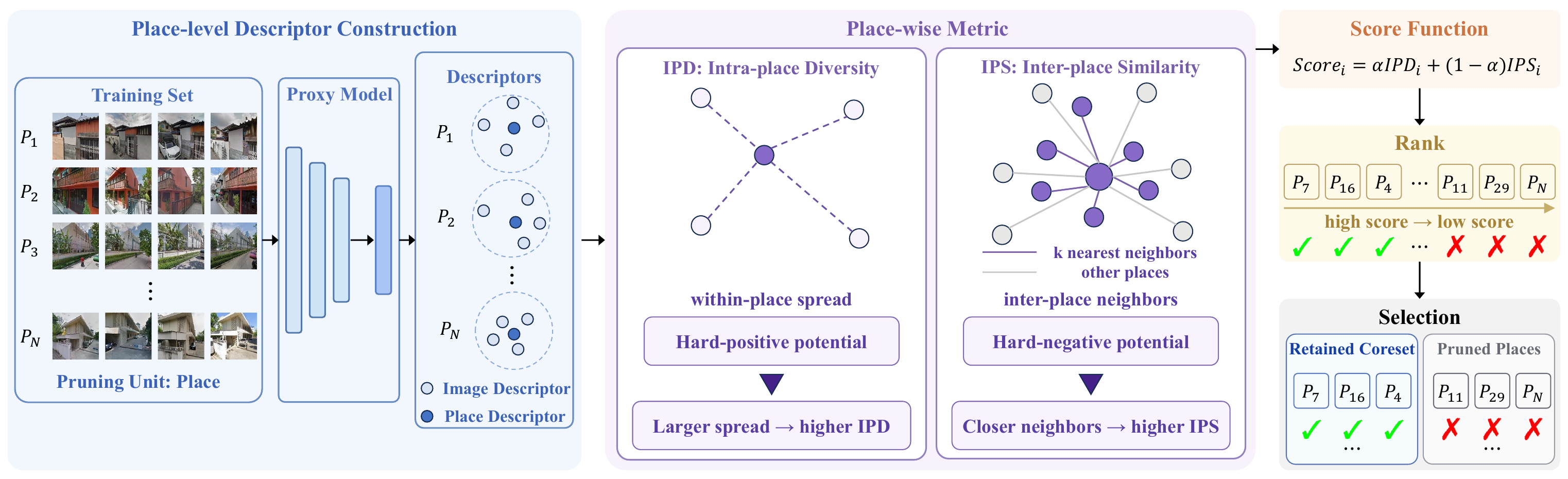}
    \vspace{-0.8cm}
    \caption{Overview of the proposed place-wise coreset selection framework. Given a place-organized training set, a proxy model is first used to extract image descriptors, which are then aggregated within each place to construct place-level descriptors. Each place is evaluated by two complementary metrics, i.e., intra-place diversity (IPD) and inter-place similarity (IPS), which respectively characterize its hard-positive and hard-negative potential. The two metrics are weighted and fused into a unified score to rank all places. Finally, high-score places are retained to form the coreset, while low-score places are pruned.}
    \label{fig:pipeline}
    \vspace{-0.5cm}
\end{figure*}

\section{Methodology}
\label{sec:methogology}
In this section, we present the proposed place-wise dataset pruning framework tailored for VPR, as shown in Fig. \ref{fig:pipeline}. We first formulate dataset pruning as a place-level coreset selection problem and construct place-level descriptors using a proxy model. We then introduce two novel metrics, intra-place diversity (IPD) and inter-place similarity (IPS), which evaluate the training value of each place from two complementary perspectives. Finally, we describe a scalable selection strategy that fuses the two metrics to obtain the final coreset.

\subsection{Problem Formulation and Place-wise Pruning}
Consider a full training dataset, e.g., GSV-Cities \cite{gsv_cities}, denoted as $\mathcal{D}=\{P_1,P_2,\dots,P_N\}$, where each $P_i$ corresponds to a place. Each place is depicted by a set of $K_i$ images, generally captured under different viewpoints and conditions, and is denoted as $P_i=(\{I^i_1,I^i_2,\dots,I^i_{K_i}\},y_i)$, where $y_i$ is the place label. Our goal is to select an informative and compact subset $\mathcal{S} \subset \mathcal{D}$, such that the performance gap between models trained on $\mathcal{S}$ and on $\mathcal{D}$ is as small as possible. The selected subset, also known as the coreset, is denoted as $\mathcal{S}=\{\hat{P}_1,\hat{P}_2,\dots,\hat{P}_M\}$. The pruning ratio $r$ is defined as $r = 1 - \frac{M}{N}$, indicating the proportion of places removed from $\mathcal{D}$. Unlike sample-wise pruning in image classification, our pruning is performed in a place-wise manner. That is, each place is retained or discarded as a whole. All images from retained places are used for training, while all images from pruned places are removed.

\subsection{Place-level Descriptor Construction}
To evaluate the training value of each place, we first obtain image-level global descriptors and then construct place-level descriptors. In VPR, each image is commonly encoded into a global descriptor that captures its holistic visual characteristics. The dominant paradigm follows a backbone-plus-aggregator architecture. Specifically, given an image $I\in\mathbb{R}^{h\times w\times c}$, a backbone network first extracts local or patch-level features, which are then aggregated into a global descriptor by an aggregation module (e.g., NetVLAD \cite{netvlad}). In our pruning framework, a proxy VPR model $\Phi(\cdot)$ is used to extract such descriptors for all training images.

For an image $I^i_j$ from $P_i$, its global descriptor is denoted as
\begin{equation}
    \mathbf{f}^i_j = \Phi(I^i_j), \quad \mathbf{f}^i_j \in \mathbb{R}^{d},
    \label{eq:image-descriptor}
\end{equation}
where $d$ denotes the descriptor dimension. Here, $\mathbf{f}^i_j$ is already $\ell_2$-normalized, following common VPR practice.

Since pruning is performed at the place level, image-level descriptors are further aggregated within each place. For place $P_i$ with $K_i$ images, we first compute its mean descriptor as
\begin{equation}
    \mathbf{m}^i = \frac{1}{K_i}\sum_{j=1}^{K_i} \mathbf{f}^i_j.
    \label{eq:mean-feature}
\end{equation}
After averaging, we further apply $\ell_2$ normalization to obtain the place-level descriptor $\bar{\mathbf{f}}^i$, which serves as the prototype representation of place $P_i$. The image descriptors $\{\mathbf{f}^i_j\}_{j=1}^{K_i}$ are used to measure intra-place variation, while the place-level descriptors $\{\bar{\mathbf{f}}^i\}_{i=1}^{N}$ are used to estimate inter-place similarity.

Following common practice in dataset pruning \cite{moderate_ds,fdmat,d2pruning}, the proxy model is used only to extract image descriptors for subsequent place-wise scoring. Our framework is not restricted to a specific proxy model, and the influence of different proxy models is further analyzed in Sec. \ref{sec:ablation_study}.

\begin{figure*}[t]
    \centering
    \vspace{-0.3cm}
    \includegraphics[width=1.0\linewidth]{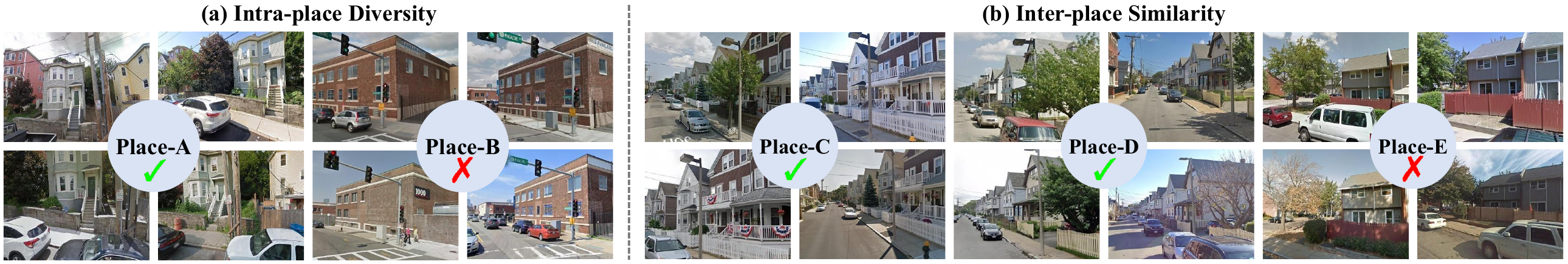}
    \vspace{-0.8cm}
    \caption{Illustration of the proposed IPD and IPS metrics. (a) IPD measures the appearance variation within the same place. Place-A exhibits larger viewpoint and appearance changes than Place-B, indicating a stronger potential to provide informative hard positive pairs. (b) IPS measures the visual proximity between different places. Place-C and Place-D are visually similar and can form hard negative pairs, while Place-E is relatively less similar to its neighboring places. Places with high IPD or high IPS are therefore assigned higher scores and are more likely to be retained in the final coreset.}
    \label{fig:ipd&ips}
    \vspace{-0.3cm}
\end{figure*}

\subsection{Place-wise Metric Design}
In VPR training, samples are inherently interdependent because the learning objective is usually defined over image pairs rather than isolated images. Images from the same place form positive pairs, while images from different places serve as negatives. Therefore, an informative place should provide useful pairwise supervision from two aspects: sufficient visual variation within the place for informative positive pairs, and high visual similarity to other places for hard negative pairs. Motivated by this observation, we design two place-wise metrics, namely intra-place diversity (IPD) and inter-place similarity (IPS), to estimate the training value of each place.

\textbf{Intra-Place Diversity.}
For place $P_i$, its image descriptors are denoted as $\{\mathbf{f}^i_1,\mathbf{f}^i_2,\dots,\mathbf{f}^i_{K_i}\}$, and its place-level descriptor is denoted as $\bar{\mathbf{f}}^i$. To measure the visual variation within the place, we define IPD as the average deviation between image descriptors and the corresponding place descriptor:
\begin{equation}
    \text{IPD}_i =
    \frac{1}{K_i}
    \sum_{j=1}^{K_i}
    \|
    \mathbf{f}^i_j - \bar{\mathbf{f}}^i
    \|_2 .
    \label{eq:ipd}
\end{equation}
A higher $\text{IPD}_i$ indicates that images within $P_i$ exhibit larger appearance changes, such as viewpoint, illumination, or condition variations. Such places are more likely to provide informative positive pairs during training. As illustrated in Fig. \ref{fig:ipd&ips} (a), Place-A has larger intra-place variation than Place-B and is therefore expected to be more valuable for learning viewpoint- and condition-robust representations.

\textbf{Inter-Place Similarity.}
While IPD focuses on positive pairs, negative pairs across different places are also essential for learning discriminative representations. Places that are visually similar to other places tend to provide harder negatives and thus help the model distinguish between perceptually similar scenes. Given the place-level descriptors $\{\bar{\mathbf{f}}^1,\bar{\mathbf{f}}^2,\dots,\bar{\mathbf{f}}^N\}$, we compute the cosine similarity between two places as
\begin{equation}
    \mathbf{S}_{ij} =
    \frac{
    \langle \bar{\mathbf{f}}^i, \bar{\mathbf{f}}^j \rangle
    }{
    \|\bar{\mathbf{f}}^i\|_2
    \|\bar{\mathbf{f}}^j\|_2
    },
    \quad i,j \in {1,\dots,N}.
    \label{eq:cos_similarity}
\end{equation}
Since place-level descriptors are $\ell_2$-normalized, this similarity is equivalent to the inner product. For each place $P_i$, we retrieve its $k$ nearest neighboring places according to $\mathbf{S}{ij}$ and define the inter-place similarity as
\begin{equation} 
    \text{IPS}_i = \frac{1}{k} \sum_{j} \mathbf{S}_{ij}, 
    \quad j \in \mathcal{N}i^k, 
    \label{eq:ips} 
\end{equation}
where $\mathcal{N}^k_i$ denotes the index set of the $k$ nearest neighboring places of $P_i$. A higher $\text{IPS}_i$ indicates that $P_i$ lies in a visually crowded region of the descriptor space, suggesting a stronger potential to form hard negative pairs with nearby places. As shown in Fig. \ref{fig:ipd&ips} (b), visually similar places tend to have higher IPS values and are therefore more likely to be retained.

\subsection{Scalable Coreset Selection}
After obtaining the IPD and IPS scores, we combine them to rank places and construct the final coreset. A straightforward implementation of IPS requires computing pairwise similarities among all place-level descriptors, which has quadratic complexity with respect to the number of places. This becomes computationally expensive for large-scale VPR datasets, where the number of places can reach tens or hundreds of thousands. To improve scalability, we simply divide all places into sequential selection mini-batches, each containing at most $B$ places, without any additional clustering or sampling strategy.
 For each mini-batch, IPD and IPS are computed for the places inside it, and IPS is estimated using the $k$ nearest neighboring places within the same mini-batch.

Since IPD and IPS may have different numerical ranges, we first apply min--max normalization to each metric within the mini-batch to align their scales. The normalized scores are denoted as $\widetilde{\text{IPD}}_i$ and $\widetilde{\text{IPS}}_i$, respectively. The final score of place $P_i$ is then computed as
\begin{equation}
    \text{Score}_i =
    \alpha \widetilde{\text{IPD}}_i
    +
    (1-\alpha)\widetilde{\text{IPS}}_i ,
    \label{eq:final-score}
\end{equation}
where $\alpha$ is a balancing coefficient controlling the relative contributions of IPD and IPS. A larger score suggests that the place has higher training value, as it tends to provide more informative positive pairs or hard negative pairs for training.

Given a pruning ratio, places in each mini-batch are ranked according to $\text{Score}_i$, and the corresponding proportion of lowest-scoring places is removed. The retained places from all mini-batches are then merged to form the final coreset. This mini-batch strategy avoids full pairwise comparison over all places, thereby improving the efficiency and scalability of coreset selection on large-scale VPR datasets. The overall procedure is summarized in Algorithm \ref{alg:place-wise-selection}.

\begin{algorithm}[t]
\caption{Place-wise Coreset Selection}
\label{alg:place-wise-selection}
    \begin{algorithmic}[1]
    \REQUIRE Training set $\mathcal{D}$, proxy model $\Phi$, pruning ratio $r$, mini-batch size $B$, neighbors $k$, weight $\alpha$
    \ENSURE Selected coreset $\mathcal{S}$
    \STATE Extract image descriptors \& construct place descriptors.
    \STATE Partition all places into selection mini-batches.
    \FOR{each mini-batch}
        \STATE Compute $\text{IPD}$ and $\text{IPS}$ for all places.
        \STATE Normalize the two metrics and compute scores.
        \STATE Retain the top $(1-r)\times B$ places with higher scores.
    \ENDFOR
    \STATE Merge all the retained places to obtain $\mathcal{S}$.
    \RETURN $\mathcal{S}$
    \end{algorithmic}
\end{algorithm}

\section{Experiments}
\label{sec:experiments}
\subsection{Datasets \& Evaluation Metrics}
\textbf{Datasets.} We evaluate our method on several benchmark datasets encompassing diverse real-world challenges. Pitts30k \cite{pitts} primarily features viewpoint variants. MSLS \cite{msls} (including MSLS-val and MSLS-challenge) contains images from urban and suburban environments under various visual changes. Nordland \cite{nordland}, captured by a front-mounted train camera, spans all four seasons. A summary is provided in Table \ref{tab: description of datasets}. \textit{More details are in the supplementary material.}

\textbf{Evaluation Metrics.} Following prior studies \cite{netvlad,vprbenchmark2}, we use the Recall@N (R@N) for model performance evaluation. Specifically, R@N measures the percentage of query images for which at least one of the top-N retrieved candidates lies within a predefined ground-truth threshold. The threshold is set to $\pm10$ frames for Nordland and 25 meters for others, consistent with other methods \cite{salad,boq,effovpr,supervlad}.

\subsection{Implementation Details}
All experiments are implemented on NVIDIA RTX A6000 GPUs. Unless otherwise specified, we use DINOv2-B \cite{dinov2} as the backbone, with only the last four transformer blocks fine-tuned and the remaining layers frozen. A NetVLAD aggregator \cite{netvlad} with 8 clusters is appended to produce 6144-dimensional global descriptors. We use GSV-Cities \cite{gsv_cities} as the source training dataset, which contains 62,514 places. Under each pruning ratio, GSV-Cities is first pruned to obtain the corresponding coreset, on which the final VPR model is trained with the multi-similarity loss \cite{msloss}. We use the Adam optimizer with an initial learning rate of $5\times10^{-5}$, halved every 3 epochs. Early stopping is applied if the performance on MSLS-val does not improve for 6 epochs. Input images are resized to $224\times224$ during training and $322\times322$ during inference, following recent VPR practice \cite{salad,boq,supervlad}. For coreset selection, the publicly available SALAD model \cite{salad} is used as the default proxy model to extract image descriptors. For place-wise scoring, the balancing coefficient $\alpha$ in Eq. (\ref{eq:final-score}) is set to 0.2. Under pruning ratios of 30\%, 50\%, and 70\%, the selection mini-batch size $B$ is set to 200, 120, and 120, respectively, and the number of nearest neighbors $k$ is set to 3, 1, and 3, respectively. \textit{Additional analyses on the generalization of the selected coresets across VPR models with different backbones, results under more pruning ratios, and hyperparameters are provided in the supplementary material.}

\begin{table}[t]
    \centering
    \caption{Summary of Several VPR Benchmark Datasets.}
    \vspace{-0.2cm}
    \begin{tabular}{c | c c | c c}
    \toprule
    \multirow{2}{*}{Dataset} & \multicolumn{2}{c|}{Variations} & \multicolumn{2}{c}{Number} \\
    \cline{2-5}
    & Viewpoint & Condition & Database & Queries \\
    \hline
    Pitts30k & \cmark & \xmark & 10,000 & 6,816 \\
    MSLS-val & \cmark & \cmark & 18,871 & 740 \\
    MSLS-challenge & \cmark & \cmark & 38,770 & 27,092 \\
    Nordland & \xmark & \cmark & 27,592 & 27,592 \\
    \bottomrule
    \end{tabular}
    \label{tab: description of datasets}
    \vspace{-0.3cm}
\end{table}

\subsection{Comparison with State-of-the-Art DP Methods}
\textbf{Baselines.} We compare our method with several representative dataset pruning approaches, including random selection, four geometry-based methods, i.e., Herding \cite{herding}, K-Center \cite{k-center}, Moderate \cite{moderate_ds}, and FDMat \cite{fdmat}, and two hybrid methods, i.e., CCS \cite{ccs} and D2Pruning \cite{d2pruning}. \textit{The details of these baselines are provided in the supplementary material.} Since these methods are originally designed for sample-wise image classification, we adapt them to the place-wise VPR setting for fair comparison. Specifically, all methods perform pruning at the place level, which is further justified by the image-wise versus place-wise comparison in Sec. \ref{sec:ablation_study}. For geometry-based methods, selection is conducted based on place-level descriptors rather than individual-image features. For hybrid methods that require a difficulty score, conventional score-based criteria such as EL2N \cite{el2n} are not naturally applicable, since VPR losses are computed over image pairs rather than independent samples. Therefore, we use IPD as a VPR-compatible difficulty measure, so that hybrid baselines can be fairly evaluated under the same place-wise pruning setting. 

\textbf{Performance Analysis.} Table \ref{tab:performance_comparison_with_sota_dp} reports the comparison results under different pruning ratios. Overall, our method consistently achieves the best or comparable performance across all four benchmark datasets. When pruning 30\% of the training places, the model trained on our coreset almost matches the full-dataset baseline on Pitts30k and MSLS-val, while further improving R@1 on Nordland and MSLS-challenge by 1.0 and 1.3 percentage points, respectively. This indicates that a considerable portion of the original training set is redundant, and the proposed place-wise scoring strategy can effectively preserve informative places for VPR training. As the pruning ratio increases, the advantage of our method becomes more evident. At the 50\% pruning ratio, our method still maintains nearly full-dataset performance on Pitts30k and MSLS-val, with only 0.1 and 0.4 percentage point drops in R@1, respectively. Meanwhile, it clearly outperforms the second-best D2Pruning by 0.4, 0.7, 5.0, and 0.5 percentage points in R@1 on Pitts30k, MSLS-val, Nordland, and MSLS-challenge, respectively. Under the more aggressive 70\% pruning ratio, our method continues to achieve the best results on all datasets. In particular, on Nordland, our method obtains 84.3\% R@1, outperforming D2Pruning by 9.1 percentage points and random selection by 16.7 percentage points. These results demonstrate that the proposed IPD- and IPS-guided selection can construct compact yet informative coresets, preserving both robustness and discriminative capability under different pruning ratios.

\begin{table*}[t]
    \centering
    \vspace{-0.3cm}
    \caption{Comparison to State-of-the-Art DP Methods on Four VPR Benchmark Datasets with Varying Pruning Ratios. The Best Results Are Highlighted in \textbf{bold}. A 0\% Pruning Ratio Indicates that We Use the Full Dataset for Training.}
    \vspace{-0.2cm}
    \setlength{\tabcolsep}{2.5mm}
    \begin{tabular}{c|c|ccc|ccc|ccc|ccc}
        \toprule
        \multirow{2}{*}{Pruning Ratio} & \multirow{2}{*}{Method} & \multicolumn{3}{c|}{Pitts30k} & \multicolumn{3}{c|}{MSLS-val} & \multicolumn{3}{c|}{Nordland} & \multicolumn{3}{c}{MSLS-challenge} \\
        \cline{3-14}
        & & R@1 & R@5 & R@10 & R@1 & R@5 & R@10 & R@1 & R@5 & R@10 & R@1 & R@5 & R@10 \\
        \hline
        \rowcolor{gray!15} \cellcolor{white} 0\% & Full Dataset & 92.9 & 97.0 & 97.9 & 92.6 & 96.4 & 97.0 & 87.3 & 94.4 & 96.3 & 76.1 & 88.1 & 91.0 \\
        \hline
        \multirow{8}{*}{30\%} & Random & 92.3 & 96.5 & 97.5 & 90.9 & 95.9 & 96.5 & 75.6 & 86.3 & 89.1 & 73.4 & 84.9 & 88.5 \\
        & Herding & 92.1 & 96.6 & 97.7 & 90.5 & 95.8 & 96.2 & 76.7 & 88.0 & 91.5 & 72.5 & 85.4 & 87.8 \\
        & K-Center & 91.9 & 96.7 & 97.7 & 91.3 & 96.3 & 96.6 & 75.3 & 87.1 & 90.4 & 73.0 & 85.0 & 88.3 \\
        & Moderate & 92.3 & 96.6 & 97.7 & 91.5 & 96.1 & 96.7 & 79.8 & 90.4 & 93.5 & 71.2 & 84.6 & 88.7 \\
        & CCS & 92.0 & 96.4 & 97.6 & 91.4 & 96.2 & 96.5 & 77.1 & 88.3 & 91.7 & 74.8 & 86.5 & 89.3 \\
        & FDMat & 92.2 & 96.5 & 97.6 & 90.4 & 95.9 & 96.5 & 77.9 & 88.7 & 92.1 & 73.3 & 85.4 & 88.2 \\
        & D2Pruning & 92.8 & 96.7 & 97.7 & 91.9 & 96.2 & 96.6 & 84.8 & 93.2 & 95.5 & 77.0 & 88.3 & 90.5 \\
        \rowcolor{lightblue} \cellcolor{white} & Ours & \textbf{92.9} & \textbf{96.8} & \textbf{97.8} & \textbf{92.6} & \textbf{96.4} & \textbf{96.8} & \textbf{88.3} & \textbf{95.2} & \textbf{97.0} & \textbf{77.4} & \textbf{88.5} & \textbf{91.2} \\
        \hline
        \hline
        \multirow{8}{*}{50\%} & Random & 91.9 & 96.4 & 97.3 & 90.1 & 95.6 & 96.4 & 71.1 & 83.2 & 87.4 & 71.4 & 84.9 & 88.3 \\
        & Herding & 91.7 & 96.5 & 97.5 & 89.7 & 95.7 & 96.1 & 73.7 & 85.5 & 89.4 & 72.2 & 85.0 & 87.4 \\
        & K-Center & 91.7 & 96.4 & 97.5 & 89.6 & 96.2 & 96.5 & 73.3 & 85.2 & 88.9 & 71.4 & 84.2 & 87.2 \\
        & Moderate & 91.8 & 96.4 & 97.5 & 90.8 & 95.8 & 96.5 & 73.6 & 85.5 & 89.5 & 70.5 & 84.6 & 88.2 \\
        & CCS & 91.7 & 96.3 & 97.3 & 90.8 & 95.7 & 96.4 & 75.8 & 87.0 & 90.6 & 70.5 & 84.6 & 88.2 \\
        & FDMat & 92.0 & 96.4 & 97.4 & 89.6 & 95.5 & 95.9 & 75.3 & 87.3 & 90.9 & 71.4 & 85.3 & 88.2 \\
        & D2Pruning & 92.4 & 96.5 & \textbf{97.7} & 91.5 & 96.0 & 96.5 & 80.6 & 90.2 & 93.0 & 75.5 & \textbf{87.1} & 89.1 \\
        \rowcolor{lightblue} \cellcolor{white} & Ours & \textbf{92.8} & \textbf{96.8} & \textbf{97.7} & \textbf{92.2} & \textbf{96.4} & \textbf{96.8} & \textbf{85.6} & \textbf{93.9} & \textbf{96.0} & \textbf{76.0} & \textbf{87.1} & \textbf{90.1} \\
        \hline
        \hline
        \multirow{8}{*}{70\%} & Random & 91.5 & 96.2 & 97.1 & 89.9 & 95.4 & 96.1 & 67.6 & 80.1 & 84.3 & 71.2 & 84.0 & 86.9 \\
        & Herding & 91.7 & 96.3 & 97.5 & 89.5 & 95.1 & 95.9 & 68.8 & 81.5 & 86.2 & 70.5 & 84.7 & 87.3 \\ 
        & K-Center & 91.5 & 96.1 & 97.3 & 88.6 & 95.3 & 96.1 & 65.0 & 79.3 & 84.8 & 69.8 & 82.8 & 85.9 \\ 
        & Moderate & 91.6 & 96.3 & 97.4 & 89.6 & 95.4 & 96.3 & 67.3 & 81.0 & 85.9 & 70.1 & 84.2 & 87.3 \\
        & CCS & 91.6 & 96.2 & 97.2 & 89.9 & 95.1 & 96.2 & 68.0 & 81.7 & 86.9 & 71.6 & 84.7 & 87.6 \\
        & FDMat & 91.7 & 96.3 & 97.4 & 89.5 & 95.1 & 96.1 & 71.9 & 84.7 & 89.6 & 71.7 & 84.8 & 87.7 \\
        & D2Pruning & 92.2 & 96.3 & 97.5 & 90.3 & 95.7 & 96.3 & 75.2 & 87.0 & 89.7 & 74.3 & 85.9 & 88.8 \\
        \rowcolor{lightblue} \cellcolor{white} & Ours & \textbf{92.5} & \textbf{96.6} & \textbf{97.6} & \textbf{91.0} & \textbf{96.0} & \textbf{96.6} & \textbf{84.3} & \textbf{93.2} & \textbf{95.7} & \textbf{75.9} & \textbf{86.9} & \textbf{89.9} \\
        \bottomrule
    \end{tabular}
    \label{tab:performance_comparison_with_sota_dp}
    \vspace{-0.5cm}
\end{table*}

\begin{table}[t]
    \centering
    \caption{Application on a Larger-Scale Dataset (i.e., the Merged Dataset). $\dagger$ Our Coreset Is Obtained by Pruning the Merged Dataset to Approximately Match the Size of GSV-Cities Using Our Method (Pruning Ratio: 71\%), and then Used for Training.}
    \vspace{-0.2cm}
    \setlength{\tabcolsep}{1.5mm}
    \begin{tabular}{l|c|cc|cc|cc}
        \toprule
        \multirow{2}{*}{Method} & Training & \multicolumn{2}{c|}{Pitts30k} & \multicolumn{2}{c|}{MSLS-val} & \multicolumn{2}{c}{Nordland} \\ 
        \cline{3-8}
        & Dataset & R@1 & R@5 & R@1 & R@5 & R@1 & R@5 \\
        \hline
        NetVLAD & \multirow{3}{*}{Merged} & 93.4 & 97.1 & 95.1 & 97.4 & 96.5 & 98.8 \\
        SALAD &  & 93.1 & 96.9 & 94.6 & 97.4 & 95.3 & 98.1 \\
        BoQ &  & 93.2 & 96.8 & 94.5 & 97.2 & 95.7 & 98.5 \\
        \hline
        \hline
        NetVLAD & GSV-Cities & 92.9 & \textbf{97.0} & 92.6 & 96.4 & 87.3 & 94.4 \\
        \rowcolor{lightblue} NetVLAD$^\dagger$ & Our Coreset$^\dagger$ & \textbf{93.3} & \textbf{97.0} & \textbf{94.5} & \textbf{96.9} & \textbf{97.0} & \textbf{99.1} \\
        \hline
        SALAD & GSV-Cities & 92.5 & 96.4 & 92.2 & 96.4 & 89.7 & 95.5 \\
        \rowcolor{lightblue} SALAD$^\dagger$ & Our Coreset$^\dagger$ & \textbf{93.1} & \textbf{96.7} & \textbf{94.3} & \textbf{97.2} & \textbf{96.1} & \textbf{98.7} \\
        \hline
        BoQ & GSV-Cities & 92.5 & 96.5 & 92.2 & 96.4 & 84.8 & 93.3 \\
        \rowcolor{lightblue} BoQ$^\dagger$ & Our Coreset$^\dagger$ & \textbf{93.1} & \textbf{96.8} & \textbf{94.1} & \textbf{96.9} & \textbf{95.8} & \textbf{98.4} \\
        \bottomrule
    \end{tabular}
    \label{tab:application_on_merged_dataset}
    \vspace{-0.5cm}
\end{table}

\subsection{Application on a Larger-Scale Merged Dataset}
Although GSV-Cities has become a widely used training dataset for recent VPR methods, larger-scale training data have been shown to further improve model performance. SelaVPR++ \cite{selavpr++} reconstructs Pitts30k-train, MSLS-train, and a subset of SF-XL into the same format as GSV-Cities, forming a merged dataset with over 210,000 places. However, training on such a large dataset also introduces substantial storage and computational costs.

To evaluate the scalability and practicality of the proposed method, we apply our place-wise pruning strategy to the merged dataset. Specifically, we prune the merged dataset by 71\%, resulting in a coreset with a scale comparable to GSV-Cities. We use DINOv2-B as the backbone and evaluate the selected coreset with three representative VPR aggregators, i.e., NetVLAD, SALAD, and BoQ. The image resolution is set to $224\times224$ during training and $322\times322$ during inference. As shown in Table \ref{tab:application_on_merged_dataset}, our coreset achieves competitive performance compared with models trained on the full merged dataset, while using only 29\% of the training places. For example, with NetVLAD, the coreset-trained model achieves 94.5\% R@1 on MSLS-val and 97.0\% R@1 on Nordland, which is close to or even higher than the model trained on the full merged dataset. More importantly, when compared with models trained on GSV-Cities, which has a similar scale to our coreset, our method consistently achieves better performance across all aggregators and test datasets. With NetVLAD, our coreset improves R@1 by 1.9 and 9.7 percentage points on MSLS-val and Nordland, respectively. These results demonstrate that our method can effectively select informative places from large-scale datasets, constructing a compact yet powerful coreset while substantially reducing the training data size.

\begin{table}[t]
    \centering
    \caption{Comparison to Multi-Dataset Trained SOTA VPR Methods.}
    \vspace{-0.2cm}
    \setlength{\tabcolsep}{0.4mm}
    \begin{tabular}{c c | c c | c c | c c | c c}
        \toprule
        \multirow{2}{*}{Method} & \multirow{2}{*}{Training Set} & \multicolumn{2}{c|}{Pitts30k} & \multicolumn{2}{c|}{MSLS-val} & \multicolumn{2}{c|}{Nordland} & 
        \multicolumn{2}{c}{Average} \\
        \cline{3-10}
        & & R@1 & R@5 & R@1 & R@5 & R@1 & R@5 & R@1 & R@5 \\
        \hline
        SALAD-CM & GSV+MSLS & 92.7 & 96.8 & 94.2 & \textbf{97.2} & 96.0 & 98.5 & 94.3 & 97.5 \\
        SelaVPR++ & Merged Dataset & 93.3 & 96.6 & \textbf{94.5} & 96.9 & 94.6 & 97.8 & 94.1 & 97.1 \\
        MegaLoc & MegaLocSet & \textbf{94.1} & \textbf{97.2} & 93.5 & 96.9 & 94.4 & 97.9 & 94.0 & 97.3 \\
        \rowcolor{lightblue} Ours & Our Coreset & 93.3 & 97.0 & \textbf{94.5} & 96.9 & \textbf{97.0} & \textbf{99.1} & \textbf{94.9} & \textbf{97.7} \\
        \bottomrule
    \end{tabular}
    \label{tab:comparison_with_SOTA_VPR_models}
    \vspace{-0.5cm}
\end{table}

\subsection{Comparison with State-of-the-Art VPR Methods}
By applying the proposed pruning strategy to the merged dataset, we obtain a compact coreset with a scale comparable to GSV-Cities. Although much smaller than the original multi-dataset training set, the selected coreset can still support the training of highly competitive VPR models. To evaluate its practical value, we compare the model trained on our coreset with several state-of-the-art VPR methods using the same DINOv2-B backbone, including SALAD-CM \cite{salad-cm}, SelaVPR++ \cite{selavpr++}, and MegaLoc \cite{megaloc}. These methods rely on large-scale or carefully constructed multi-dataset training sets, while our model is trained only on the selected coreset.

As shown in Table \ref{tab:comparison_with_SOTA_VPR_models}, the model trained on our coreset achieves competitive performance against these multi-dataset trained methods and obtains the best average results, with 94.9\% R@1 and 97.7\% R@5. In particular, our model achieves the best results on Nordland and strong performance on MSLS-val, showing that the selected coreset preserves informative and generalizable places from the original large-scale training data. These results indicate that the proposed method can substantially reduce the training data size while maintaining strong VPR performance. Rather than replacing existing large-scale training paradigms, our coreset provides a compact and reusable training subset that can facilitate efficient model development and evaluation under reduced storage and computational costs.

\begin{figure}[t]
    \centering
    \includegraphics[width=1.0\linewidth]{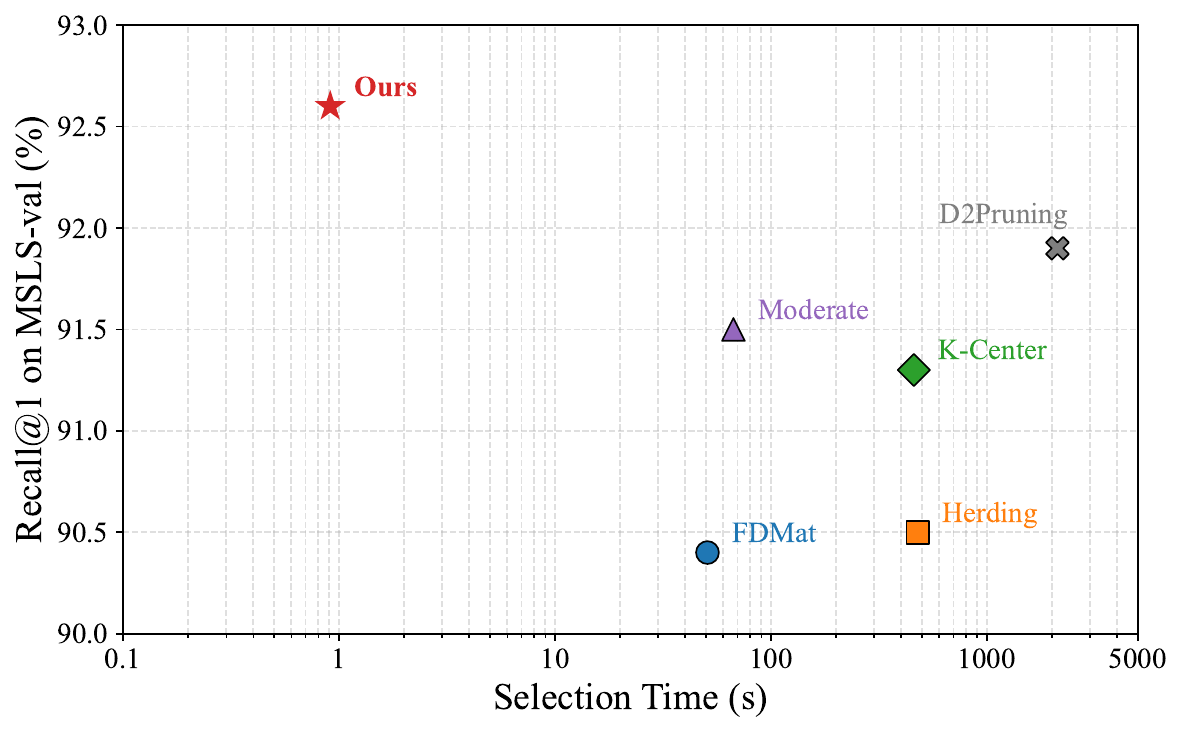}
    \vspace{-0.8cm}
    \caption{Comparison of different DP methods in method-specific selection time and Recall@1 on MSLS-val. The reported time excludes the initial feature extraction, which is shared by all methods. The pruning ratio is fixed at 30\%, and the horizontal axis is plotted on a logarithmic scale.}
    \label{fig:selection_efficiency}
    \vspace{-0.5cm}
\end{figure}

\subsection{Selection and Training Efficiency Analysis}
The primary goal of dataset pruning is to improve data and training efficiency while preserving model performance as much as possible. In this section, we analyze the efficiency of the proposed method from both the selection and training perspectives. Specifically, we first compare the selection cost of different pruning methods, then evaluate the reduction in training time brought by the selected coreset. We further report the end-to-end cost of the whole pruning-and-training pipeline and analyze the convergence behavior to show how the proposed coreset improves training efficiency.

\textbf{1) Selection Efficiency.} We first compare the method-specific selection cost of different pruning methods on GSV-Cities. \textit{The results on the merged dataset are in the supplementary material.} As shown in Fig. \ref{fig:selection_efficiency}, the pruning ratio is fixed at 30\%, and the horizontal axis reports the selection time after feature extraction. The feature extraction time is excluded here because it is shared by all methods, while the end-to-end cost (including feature extraction) will be analyzed later. Our method achieves the lowest selection time, less than 1 second, while also obtaining the best R@1 on MSLS-val. This efficiency mainly comes from the simple computation of IPD and IPS and the mini-batch-based ranking strategy, which avoids iterative subset search or full-dataset graph construction. In contrast, Herding and K-Center rely on iterative selection over place-level descriptors, resulting in non-trivial overhead. Moderate and FDMat rely on center-based statistics for selection. This requires an additional center-construction step before pruning, which introduces extra computational overhead. D2Pruning builds a graph over the dataset and performs graph-based propagation, leading to the largest overhead, about 40 minutes. These results show that the proposed method provides a better trade-off between selection efficiency and retrieval performance than existing pruning baselines.

\begin{table}[t]
    \centering
    \caption{Comparison of Total Training Time and Average Retrieval Performance Between Full Datasets and Our Coresets. Avg. R@1 Is Averaged over Pitts30k, MSLS-val, and Nordland. $\Delta T$ Denotes the Relative Training-time Change Compared with the Corresponding Full Dataset, and $\Delta$Avg. R@1 Denotes the Performance Difference.}
    \vspace{-0.2cm}
    \label{tab:training_time_efficiency}
    \setlength{\tabcolsep}{0.8mm}
    \begin{tabular}{l ccccc}
    \toprule
    Training Data & Pruning Ratio & Time (h) & $\Delta T$ & Avg. R@1 & $\Delta$Avg. R@1 \\
    \hline
    GSV-Cities & 0\% & 3.545 & -- & 90.9 & -- \\
    Our Coreset & 30\% & 2.493 & \textcolor{green}{-29.7\%} & \textbf{91.3} & \textcolor{red}{+0.4} \\
    Our Coreset & 50\% & 1.791 & \textcolor{green}{-49.5\%} & 90.2 & \textcolor{green}{-0.7} \\
    Our Coreset & 70\% & \textbf{1.576} & \textcolor{green}{-55.5\%} & 89.3 & \textcolor{green}{-1.6} \\
    \hline
    \hline
    Merged & 0\% & 19.152 & -- & \textbf{95.0} & -- \\
    Our Coreset & 71\% & \textbf{3.912} & \textcolor{green}{-79.6\%} & 94.9 & \textcolor{green}{-0.1} \\
    \bottomrule
    \end{tabular}
    \vspace{-0.3cm}
\end{table}

\textbf{2) Training Efficiency.} 
We further evaluate whether the selected coreset can reduce the actual training cost. Table \ref{tab:training_time_efficiency} compares the total training time required to obtain the best-performing model using the full datasets and our coresets. On GSV-Cities, the 30\% pruned coreset reduces the total training time from 3.545 hours to 2.493 hours, corresponding to a 29.7\% reduction, while maintaining nearly the same (even better) average R@1. Similar trends can be observed under higher pruning ratios, where the training time is further reduced with moderate performance degradation. The advantage becomes more significant on the larger merged dataset. Our 71\% pruned coreset reduces the total training time from 19.152 hours to 3.912 hours, achieving a 79.6\% reduction while keeping almost unchanged average R@1. These results show that the proposed coreset effectively improves training efficiency, especially for large-scale VPR training.

\begin{table}[t]
    \centering
    \caption{End-to-End Cost Comparison Between Full-Dataset Training and Our Pruning-and-Training pipeline. Feature Extraction and Selection Are only Required for Coreset Construction, and Their Costs are Included in the Total Time.}
    \label{tab:end_to_end_cost}
    \setlength{\tabcolsep}{0.8mm}
    \vspace{-0.2cm}
    \begin{tabular}{l c c c c c}
    \toprule
    Training Data & Feat. Ext. (h) & Selection (s) & Train (h) & Total (h) $\downarrow$ & $\Delta T$ \\
    \hline
    GSV-Cities & -- & -- & 3.545 & 3.545 & -- \\
    Our Coreset & 0.264 & 0.91 & 2.493 & \textbf{2.757} & \textcolor{green}{-22.2\%} \\
    \hline
    \hline
    Merged & -- & -- & 19.152 & 19.152 & -- \\
    Our Coreset & 0.913 & 1.49 & 3.912 & \textbf{4.825} & \textcolor{green}{-74.8\%} \\
    \bottomrule
    \end{tabular}
    \vspace{-0.3cm}
\end{table}

\textbf{3) End-to-End Cost.}
To provide a complete view of the pruning-and-training pipeline, we report the end-to-end cost in Table \ref{tab:end_to_end_cost}. Different from the training-time comparison above, this table additionally includes the feature extraction and selection time required for coreset construction. We focus on the pruning ratios that maintain nearly unchanged performance, i.e., 30\% pruning on GSV-Cities and 71\% pruning on the merged dataset. On GSV-Cities, our 30\% pruned coreset reduces the total cost from 3.545 hours to 2.757 hours, even after including the extra pruning overhead. The benefit becomes more significant on the larger merged dataset, where our 71\% pruned coreset reduces the total cost from 19.152 hours to 4.825 hours, corresponding to a 74.8\% reduction. These results show that the overhead introduced by feature extraction and coreset selection is small and does not offset the training-time savings brought by dataset pruning.

\begin{figure}[t]
    \centering
    \includegraphics[width=1.0\linewidth]{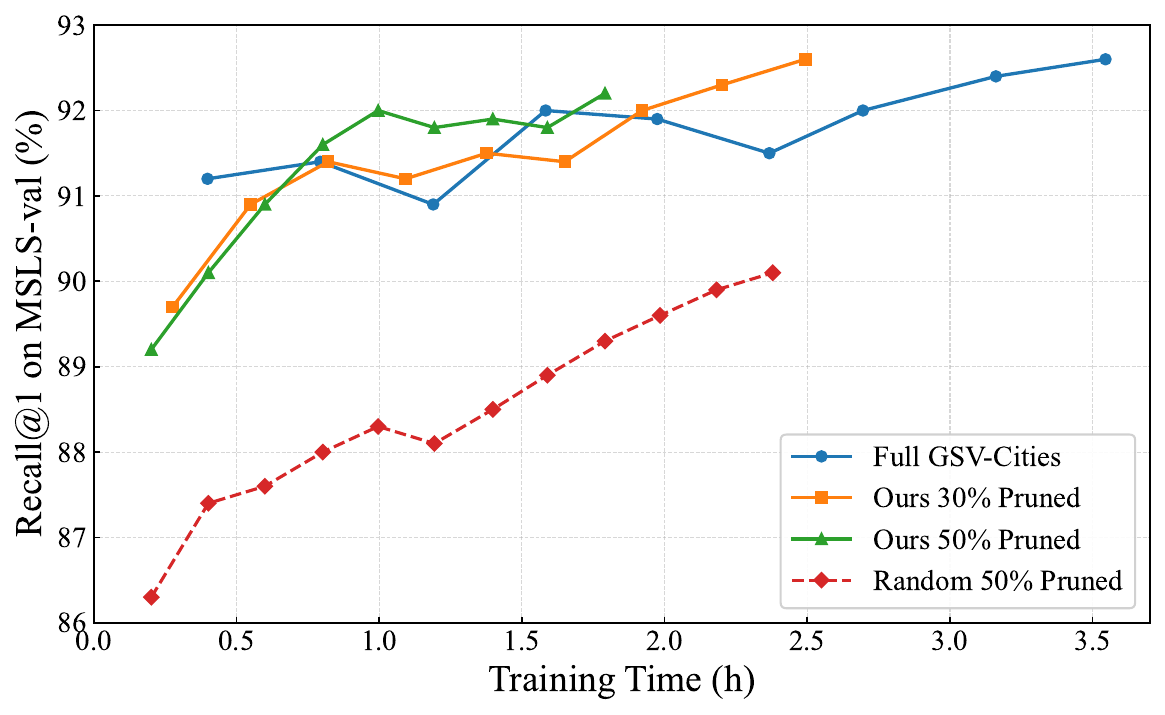}
    \vspace{-0.6cm}
    \caption{Convergence comparison on GSV-Cities. The horizontal axis denotes cumulative training time, and Recall@1 on MSLS-val is evaluated after each epoch. Each marker corresponds to one validation checkpoint.}
    \label{fig:convergence_curve}
    \vspace{-0.5cm}
\end{figure}

\textbf{4) Convergence Analysis.} We further compare the training-time convergence behavior of models trained on the full GSV-Cities dataset and different coresets. Since different training sets contain different numbers of samples, one epoch on a pruned coreset requires fewer batch updates and less training time than one epoch on the full dataset. Therefore, we use cumulative training time as the horizontal axis in Fig. \ref{fig:convergence_curve}. Compared with full-dataset training, our coresets reach competitive recognition performance within a shorter training time. In particular, the 50\% pruned coreset reaches around 92\% R@1 much earlier than the full dataset, showing that the selected subset provides effective supervision for model optimization. By contrast, random pruning under the same 50\% pruning ratio remains consistently worse throughout training, indicating that simply reducing the data volume does not guarantee efficient convergence. These results demonstrate that our selection strategy preserves more informative training places, leading to a better training-time/performance trade-off.

\subsection{Ablation Study}
\label{sec:ablation_study}
We conduct ablation studies to analyze the key design choices of the proposed framework. Specifically, we study the effect of the pruning unit, the contributions of IPD and IPS, and the influence of different proxy models. \textit{Additional hyperparameter analyses, including the effects of $\alpha$, mini-batch size $B$, and the number of nearest neighbors $k$, are provided in the supplementary material.}

\begin{table}[t]
    \centering
    \caption{Effect of Pruning Unit on GSV-Cities Under the 50\% Pruning Ratio. Image-Wise Pruning Selects Individual Images, While Place-Wise Pruning Retains/Removes Each Place as a Whole.}
    \vspace{-0.2cm}
    \label{tab:ablation_of_pruning_unit}
    \begin{tabular}{l| c| c c| c c| c c}
        \toprule
        \multirow{2}{*}{Method} & \multirow{2}{*}{Unit} & \multicolumn{2}{c|}{Pitts30k} & \multicolumn{2}{c|}{MSLS-val} & \multicolumn{2}{c}{Nordland} \\
        \cline{3-8}
        & & R@1 & R@5 & R@1 & R@5 & R@1 & R@5 \\
        \hline
        \multirow{2}{*}{Herding} & Image & 91.0 & 96.0 & 89.3 & 95.4 & 69.0 & 81.6 \\
        & Place & \textbf{91.7} & \textbf{96.5} & \textbf{89.7} & \textbf{95.7} & \textbf{73.7} & \textbf{85.5} \\
        \hline
        \multirow{2}{*}{K-Center} & Image & 90.6 & 95.8 & 88.4 & 95.1 & 67.8 & 80.3 \\
        & Place & \textbf{91.7} & \textbf{96.4} & \textbf{89.6} & \textbf{96.2} & \textbf{73.3} & \textbf{85.2} \\
        \hline
        \multirow{2}{*}{Moderate} & Image & 91.1 & 96.1 & 89.9 & 95.7 & 69.8 & 82.6 \\
        & Place & \textbf{91.8} & \textbf{96.4} & \textbf{90.8} & \textbf{95.8} & \textbf{73.6} & \textbf{85.5} \\
        \hline
        \multirow{2}{*}{FDMat} & Image & 91.4 & 96.2 & 88.7 & 95.2 & 70.6 & 83.9 \\
        & Place & \textbf{92.0} & \textbf{96.4} & \textbf{89.6} & \textbf{95.5} & \textbf{75.3} & \textbf{87.3} \\
        \bottomrule
    \end{tabular}
\end{table}

\textbf{1) Effect of Pruning Unit.} We first examine the effect of the pruning unit by comparing image-wise and place-wise pruning. To make the comparison fair and direct, we evaluate four representative methods that can be naturally applied under both settings. As shown in Table \ref{tab:ablation_of_pruning_unit}, place-wise pruning consistently outperforms its image-wise counterpart across all methods. The improvement is particularly clear on Nordland, where place-wise pruning brings 3.8--5.5 percentage point gains in R@1 over image-wise pruning. This indicates that directly pruning individual images may disrupt the relational structure of VPR training, such as weakening positive pairs within the same place or hard negative pairs across visually similar places. In contrast, place-wise pruning preserves each retained place as a complete training unit, making it better aligned with the pairwise supervision used in VPR.

\begin{table}[t]
    \centering
    \caption{Ablation of IPD and IPS on GSV-Cities Under the 50\% Pruning Ratio. Neither IPD nor IPS Corresponds to Random Selection.}
    \label{tab:ablation_ipd_ips}
    \vspace{-0.2cm}
    \setlength{\tabcolsep}{2.7mm}
    \begin{tabular}{c c | c c | c c| c c}
        \toprule
        \multirow{2}{*}{IPD} & \multirow{2}{*}{IPS} & \multicolumn{2}{c|}{Pitts30k} & \multicolumn{2}{c|}{MSLS-val} & \multicolumn{2}{c}{Nordland} \\
        \cline{3-8}
        & & R@1 & R@5 & R@1 & R@5 & R@1 & R@5 \\
        \hline
        \xmark & \xmark & 91.9 & 96.4 & 90.1 & 95.6 & 71.1 & 83.2 \\
        \cmark & \xmark & 92.3 & 96.3 & 91.9 & 96.2 & 80.6 & 90.5 \\
        \xmark & \cmark & 92.6 & 96.6 & 91.6 & 95.9 & 83.8 & 92.7 \\
        \cmark & \cmark & \textbf{92.8} & \textbf{96.8} & \textbf{92.2} & \textbf{96.4} & \textbf{85.6} & \textbf{93.9} \\
        \bottomrule
    \end{tabular}
    \vspace{-0.3cm}
\end{table}

\begin{figure*}[t]
    \centering
    \vspace{-0.3cm}
    \includegraphics[width=1.0\linewidth]{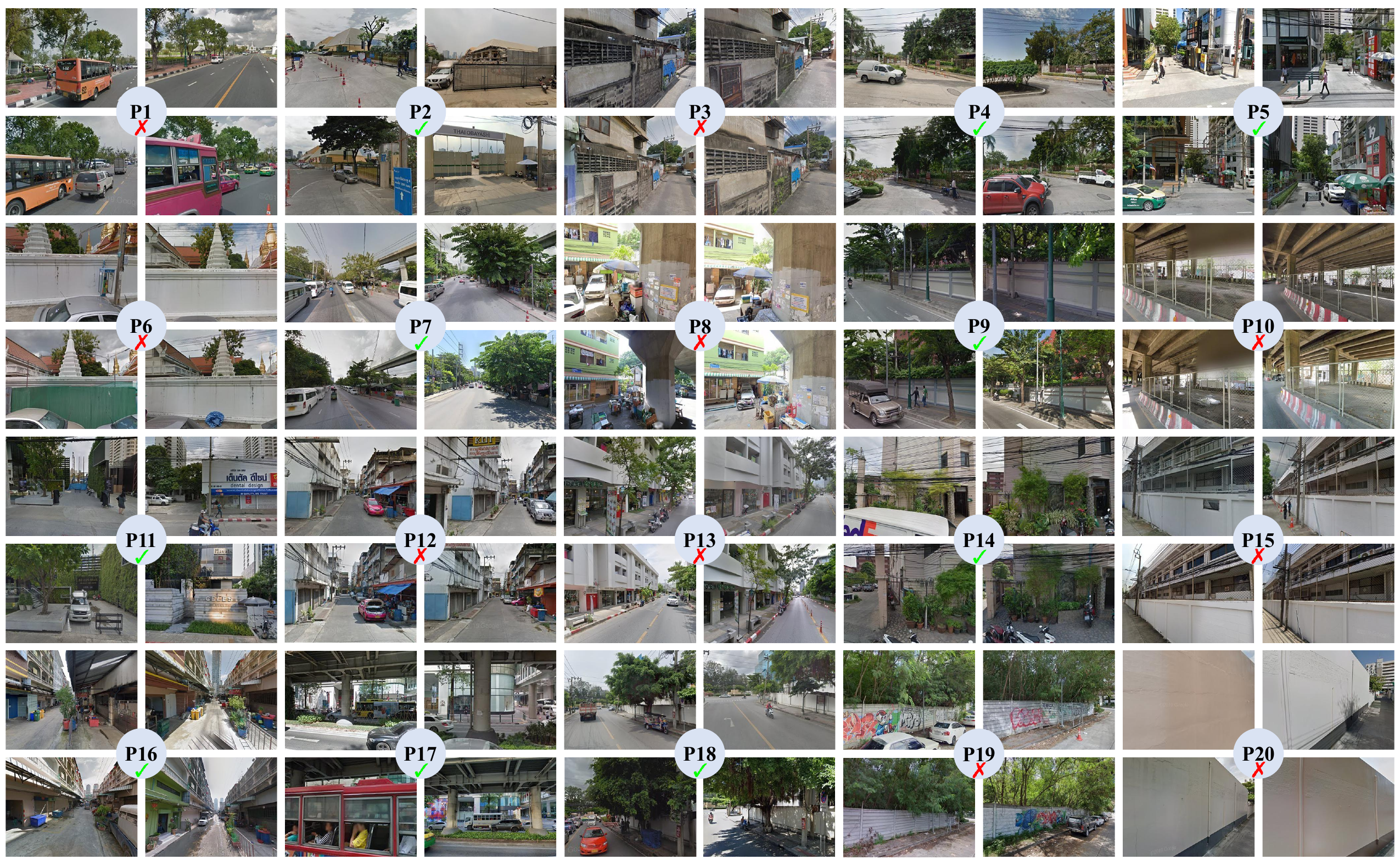}
    \vspace{-0.8cm}
    \caption{Qualitative visualization of place-wise pruning results on a randomly sampled selection mini-batch from GSV-Cities under a pruning ratio of 50\%. Each place contains four images. Places marked with a green check are retained, while those marked with a red cross are pruned.}
    \label{fig:qualitative_pruning_results}
    \vspace{-0.3cm}
\end{figure*}

\textbf{2) Effect of IPD and IPS.}
We then analyze the contributions of the two proposed metrics, i.e., intra-place diversity (IPD) and inter-place similarity (IPS). As shown in Table \ref{tab:ablation_ipd_ips}, using either IPD or IPS alone consistently improves over the random selection baseline, indicating that both intra-place variation and inter-place similarity are useful cues for evaluating the training value of places. The improvement is especially significant on Nordland, where IPD and IPS improve R@1 from 71.1\% to 80.6\% and 83.8\%, respectively. By combining IPD and IPS, our full method achieves the best performance across all datasets, reaching 92.8\%, 92.2\%, and 85.6\% R@1 on Pitts30k, MSLS-val, and Nordland. These results verify the complementary roles of IPD and IPS. IPD helps retain places with rich intra-place variations for informative positive pairs, while IPS favors visually similar places that can provide hard negative pairs. Together, they enable our method to construct a compact and informative coreset for VPR training. 

\begin{table}[t]
    \centering
    \caption{Effect of Different Proxy Models on GSV-Cities Under the 30\% Pruning Ratio. The Final VPR Model Is Fixed as DINOv2-B + NetVLAD, and only the Proxy Model Used for Coreset Selection Is Changed.}
    \setlength{\tabcolsep}{1.9mm}
    \label{tab:ablation_proxy_model}
    \vspace{-0.2cm}
    \begin{tabular}{l c c c c}
        \toprule
        Proxy Model & Pitts30k & MSLS-val & Nordland & Avg. R@1 \\
        \hline
        Pretrained DINOv2-B &92.6 & 92.4 & 87.5 & 90.8 \\
        VGG16+NetVLAD & 92.5 & 92.3 & 87.9 & 90.9 \\
        Open-source SALAD & \textbf{92.9} & \textbf{92.6} & \textbf{88.3} & \textbf{91.3} \\
        \bottomrule
    \end{tabular}
    \vspace{-0.3cm}
\end{table}

\textbf{3) Effect of Proxy Model.} We further analyze the influence of different proxy models used for descriptor extraction during coreset selection. We compare three proxy models, including pretrained DINOv2-B using the [CLS] token as the global descriptor, VGG16+NetVLAD trained on Pitts30k-train \cite{netvlad}, and the open-source SALAD model \cite{salad}. For a fair comparison, only the proxy model is changed, while the final VPR model is fixed as DINOv2-B+NetVLAD with the same training protocol. As shown in Table \ref{tab:ablation_proxy_model}, all proxy models achieve competitive performance, with the average R@1 ranging from 90.8\% to 91.3\%. This indicates that our framework is not restricted to a specific proxy architecture or training source. Among them, SALAD achieves the best overall result, suggesting that stronger VPR-oriented descriptors can provide slightly better place-wise scoring. Nevertheless, the performance gap among different proxies is small, demonstrating that our method is reasonably robust to the choice of proxy model.

\subsection{Qualitative Experiments}
To provide an intuitive view of the real selection behavior of the proposed method, we visualize the pruning results on a randomly sampled selection mini-batch under the 50\% pruning ratio as shown in Fig. \ref{fig:qualitative_pruning_results}. It can be observed that the retained places generally exhibit richer visual variations or more complex scene content. In addition, several retained places, such as P4, P7, P9 and P18, share similar layout and vegetation patterns, indicating their potential to form relatively hard negative pairs. In contrast, many pruned places appear relatively repetitive or less informative (e.g., P8, P10, P15 and P20). Although the final selection is determined by the fused score rather than by visual inspection alone, these examples suggest that the proposed place-wise scoring strategy can preserve informative and visually related places while removing relatively redundant ones.

\section{Conclusion}
\label{sec:conlusions}
In this paper, we study dataset pruning for data-efficient VPR training. Different from conventional sample-wise pruning methods, we formulate VPR dataset pruning at the place level and introduce two complementary metrics, i.e., intra-place diversity and inter-place similarity, to evaluate the training value of each place. Based on these metrics, the proposed framework constructs compact and informative coresets while reducing training and storage costs. Extensive experiments show that our method consistently outperforms existing dataset pruning approaches and requires substantially less selection time. Moreover, the proposed method scales well to a larger merged dataset and produces coresets that support efficient training across different VPR models. We believe this work lays a solid foundation for future research on scalable coreset selection for large-scale VPR datasets.

\bibliographystyle{IEEEtran}
\bibliography{IEEEabrv,main}

\appendix
\subsection{Overview}
This supplementary material provides additional experimental results and further analysis to complement the main paper. The contents are organized as follows:
\begin{itemize}
    \item Additional results under low and high pruning ratios;
    \item Additional selection-time analysis on the larger merged dataset;
    \item Hyperparameter analysis, including the effects of $\alpha$, $k$, and $B$;
    \item Generalization analysis across different backbones;
    \item Details of benchmark datasets;
    \item Details of compared dataset pruning methods;
    \item Limitations and future directions.
\end{itemize}

\subsection{Additional Results under Low and High Pruning Ratios}
In this section, we provide additional results under pruning ratios beyond those mainly discussed in the main paper. The experimental setting follows Table II in the main paper. Specifically, pruning is performed on GSV-Cities, and the open-source SALAD model is used as the proxy model to extract descriptors for coreset selection. The final VPR model is DINOv2-B with NetVLAD, trained on the selected coreset and evaluated on Pitts30k, MSLS-val, and Nordland. The only difference lies in the pruning ratios considered in this section.

\textbf{Results under Low Pruning Ratios.}
In the main paper, we observe that pruning 30\% of the training places causes almost no performance degradation in R@1 and even improves the result on Nordland. This motivates us to further examine whether mild pruning can improve the training data quality. Therefore, we evaluate our method under lower pruning ratios, i.e., 10\% and 20\%. The full dataset setting is also included for reference.

As shown in Table \ref{tab:low_pruning_ratio}, mild pruning does not degrade overall retrieval performance. Instead, pruning 10\% or 20\% of the training places slightly improves the average performance over the full dataset. In particular, the 20\% pruning setting achieves the best average R@1 and R@5, improving the average R@1 from 90.9\% to 91.5\%. This suggests that the GSV-Cities training set contains a certain amount of redundant or less informative places, and removing them can improve the overall training data quality. However, since low pruning ratios bring limited training-cost reduction, the main paper focuses more on moderate pruning ratios, where the efficiency-performance trade-off is more significant.

\textbf{Results under Extremely High Pruning Ratios.}
We further evaluate different dataset pruning methods under extremely high pruning ratios, including 90\%, 95\%, and 99\%. These settings retain only 10\%, 5\%, and 1\% of the training places, respectively, providing a challenging evaluation of coreset quality under very limited data budgets.

Table \ref{tab:high_pruning_ratio} reports the comparison results. Overall, our method consistently achieves the best performance among all competing methods under these aggressive pruning settings. The advantage is particularly clear at the 90\% and 95\% pruning ratios. For example, on Nordland, our method outperforms the second-best method by 11.8 and 5.1 percentage points in R@1 under the 90\% and 95\% pruning ratios, respectively. These results show that the proposed place-wise scoring strategy can preserve more informative training places when the retained data budget is very limited. As expected, the absolute performance of all methods decreases when the pruning ratio becomes extremely high, especially under the 99\% pruning ratio. The degradation is more pronounced on Nordland, which contains natural-scene imagery and has a larger domain gap from the predominantly urban and suburban training data. This indicates that extreme pruning may remove rare or domain-relevant places that are important for cross-domain generalization. These observations suggest that developing pruning strategies that explicitly consider scene rarity and diversity under very small data budgets may be a promising direction for future work.

\begin{table}[t]
    \centering
    \caption{Results of Lower Pruning Ratio.}
    \label{tab:low_pruning_ratio}
    \setlength{\tabcolsep}{1.3mm}
    \begin{tabular}{c | c c| c c| c c| c c}
        \toprule
        \multirow{2}{*}{Pruning Ratio} & \multicolumn{2}{c|}{Pitts30k} & \multicolumn{2}{c|}{MSLS-val} & \multicolumn{2}{c|}{Nordland} & \multicolumn{2}{c}{Average} \\
        \cline{2-9} 
        & R@1 & R@5 & R@1 & R@5 & R@1 & R@5 & R@1 & R@5 \\
        \hline
        0\% & 92.9 & \textbf{97.0} & 92.6 & 96.4 & 87.3 & 94.4 & 90.9 & 95.9 \\
        10\% & 93.0 & \textbf{97.0} & 92.6 & \textbf{96.5} & 87.7 & 94.8 & 91.1 & 96.1 \\
        20\% & \textbf{93.1} & 96.9 & \textbf{92.7} & \textbf{96.5} & \textbf{88.6} & \textbf{95.5} & \textbf{91.5} & \textbf{96.3} \\
        \bottomrule
    \end{tabular}
    \vspace{-0.3cm}
\end{table}

\begin{table}[t]
    \centering
    \caption{Comparison to Different Dataset Pruning Methods with Extremely High Pruning Ratios.}
    \label{tab:high_pruning_ratio}
    \setlength{\tabcolsep}{1.3mm}
    \begin{tabular}{c|c|cc|cc|cc}
        \toprule
        \multirow{2}{*}{Pruning Ratio} & \multirow{2}{*}{Method} & \multicolumn{2}{c|}{Pitts30k} & \multicolumn{2}{c|}{MSLS-val} & \multicolumn{2}{c}{Nordland} \\
        \cline{3-8} 
        & & R@1 & R@5 & R@1 & R@5 & R@1 & R@5 \\
        \hline
        \multirow{8}{*}{90\%} & Random & 91.1 & 95.7 & 87.6 & 94.3 & 58.8 & 72.9\\
        & Herding & 91.2 & 96.2 & 87.2 & 94.3 & 56.6 & 71.0 \\
        & K-Center & 90.7 & 96.0 & 87.3 & 94.2 & 55.1 & 70.3 \\
        & Moderate & 91.1 & 96.1 & 87.7 & 94.3 & 57.0 & 71.9 \\
        & CCS & 91.0 & 96.0 & 87.7 & 94.5 & 57.5 & 71.8 \\
        & FDMat & 91.3 & 96.1 & 87.2 & 92.8 & 61.4 & 75.9 \\
        & D2Pruning & 91.3 & 96.2 & 87.8 & 94.6 & 64.5 & 78.5 \\
        \rowcolor{lightblue} \cellcolor{white} & Ours & \textbf{91.8} & \textbf{96.5} & \textbf{89.7} & \textbf{95.3} & \textbf{76.3} & \textbf{87.8} \\
        \hline
        \multirow{8}{*}{95\%} & Random & 90.6 & 95.5 & 86.4 & 92.8 & 47.1 & 61.8 \\
        & Herding & 90.3 & \textbf{96.1} & 85.5 & 93.0 & 52.1 & 67.1 \\
        & K-Center & 90.1 & 95.8 & 85.3 & 92.6 & 51.1 & 66.5 \\
        & Moderate & 90.6 & 95.9 & 86.1 & \textbf{93.2} & 50.6 & 65.7 \\
        & CCS & 89.9 & 95.4 & 85.0 & \textbf{93.2} & 52.1 & 67.3 \\
        & FDMat & 90.9 & 96.0 & 85.0 & 92.0 & 52.1 & 66.9 \\
        & D2Pruning & 90.7 & \textbf{96.1} & 85.9 & 92.8 & 49.4 & 63.5 \\
        \rowcolor{lightblue} \cellcolor{white} & Ours & \textbf{91.0} & 96.0 & \textbf{86.6} & 92.7 & \textbf{57.2} & \textbf{72.4} \\
        \hline
        \hline
        \multirow{8}{*}{99\%} & Random & 87.9 & 94.5 & 76.4 & 87.0 & 38.7 & 53.1 \\
        & Herding & 87.9 & 94.7 & 79.1 & 88.8 & 37.9 & 52.6 \\
        & K-Center & 87.9 & 94.6 & 76.6 & 86.5 & 35.1 & 49.5 \\
        & Moderate & 88.2 & 94.5 & 79.3 & 89.1 & 37.4 & 52.4 \\
        & CCS & 88.2 & 94.7 & 77.6 & 86.2 & 37.2 & 51.6 \\
        & FDMat & 87.0 & 94.8 & 74.3 & 84.5 & 40.2 & 55.2 \\
        & D2Pruning & 87.8 & 94.4 & 74.5 & 85.7 & 38.1 & 53.5 \\
        \rowcolor{lightblue} \cellcolor{white} & Ours & \textbf{88.3} & \textbf{94.9} & \textbf{80.0} & \textbf{90.3} & \textbf{40.9} & \textbf{55.6} \\
        \bottomrule
    \end{tabular}
    \vspace{-0.3cm}
\end{table}

\begin{figure}[t]
    \centering
    \includegraphics[width=1.0\linewidth]{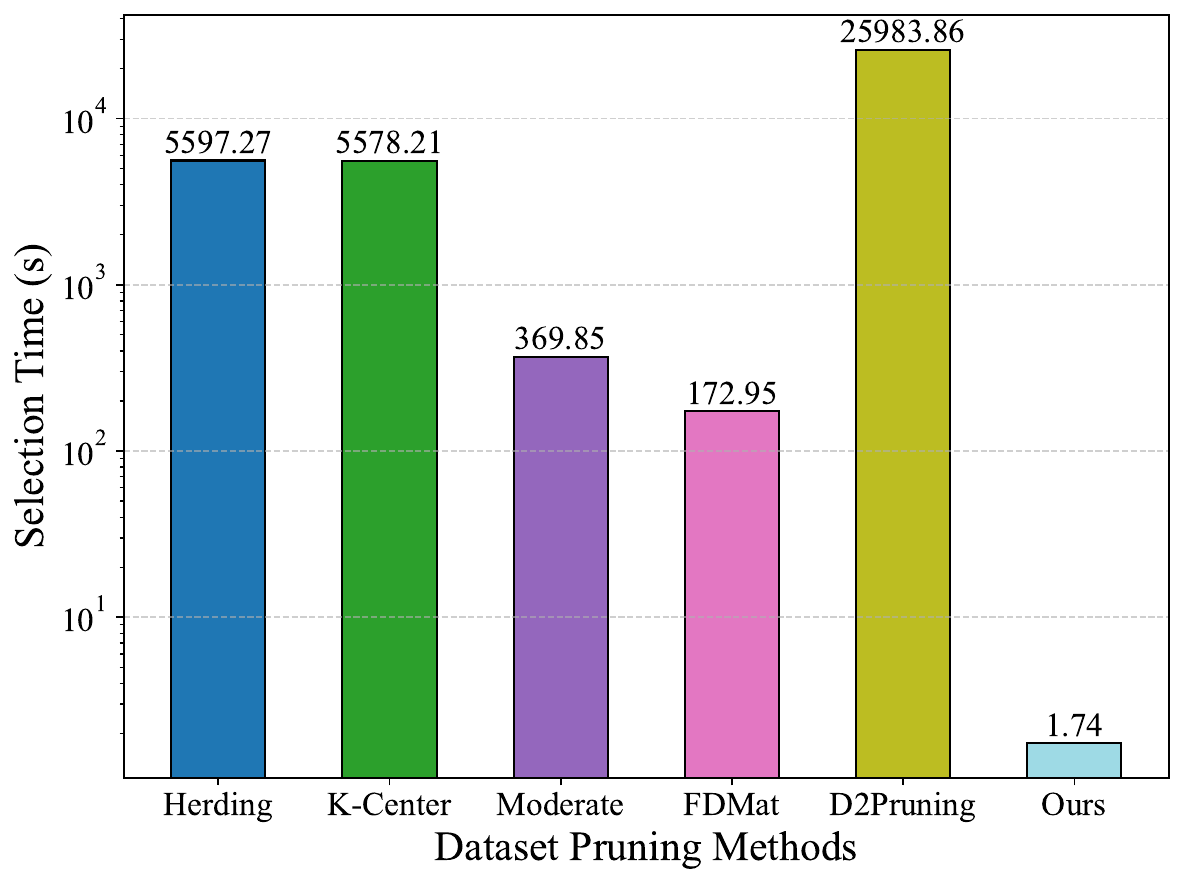}
    \vspace{-0.6cm}
    \caption{Comparison of various dataset pruning methods in selection time (seconds) on the merged dataset. The reported time excludes the initial feature extraction, which is shared by all. The pruning ratio is set to 30\%. Note that the vertical axis is plotted on a logarithmic scale.}
    \label{fig:additional_selection_time}
    \vspace{-0.3cm}
\end{figure}

\subsection{Additional Selection-Time Analysis on the Merged Dataset}
In the main paper, we report the method-specific selection time on GSV-Cities (i.e., Fig. 4). To further examine the scalability of our proposed method, we additionally evaluate the selection time on the larger merged dataset, which contains about $3.5\times$ more places than GSV-Cities. The pruning ratio is fixed at 30\%, i.e., 70\% of the training places are retained. As in the main paper, the reported time excludes the initial feature extraction, which is shared by all methods.

As shown in Fig. \ref{fig:additional_selection_time}, the selection time of several baselines increases rapidly on the larger dataset. For example, Herding requires 477.98 seconds on GSV-Cities but 5597.27 seconds on the merged dataset, showing a more than tenfold increase. K-Center shows a similar trend, and D2Pruning requires 25983.86 seconds due to graph construction and message passing over the whole dataset, making selection itself a substantial computational bottleneck. In contrast, our method only requires 1.74 seconds on the merged dataset. This is because the proposed mini-batch selection strategy avoids iterative subset search and full-dataset graph construction. These results further demonstrate the scalability and efficiency of our method on large-scale VPR datasets.

\begin{table}[t]
    \centering
    \caption{Ablation on the Weighting Coefficient $\boldsymbol{\alpha}$.}
    \setlength{\tabcolsep}{2.0mm}
    \label{tab:ablation_on_alpha}
    \vspace{-0.2cm}
    \begin{tabular}{c|cc|cc|cc|cc}
        \toprule
        \multirow{2}{*}{$\alpha$} & \multicolumn{2}{c|}{Pitts30k} & \multicolumn{2}{c|}{MSLS-val} & \multicolumn{2}{c|}{Nordland} & \multicolumn{2}{c}{Average} \\
        \cline{2-9}
        & R@1 & R@5 & R@1 & R@5 & R@1 & R@5 & R@1 & R@5 \\
        \hline
        0.1 & 92.7 & \textbf{96.8} & \textbf{92.7} & 96.2 & 88.1 & 95.0 & 91.2 & 96.0 \\
        \rowcolor{lightblue} 0.2 & \textbf{92.9} & \textbf{96.8} & 92.6 & \textbf{96.4} & \textbf{88.3} & \textbf{95.2} & \textbf{91.3} & \textbf{96.1} \\
        0.3 & \textbf{92.9} & 96.7 & 92.3 & 96.1 & 88.2 & 95.1 & 91.1 & 96.0 \\
        0.4 & 92.7 & 96.7 & 92.5 & 96.2 & 87.9 & 94.9 & 91.0 & 95.9 \\
        0.5 & 92.8 & 96.7 & 92.2 & 96.1 & 87.4 & 94.5 & 90.8 & 95.8 \\
        \bottomrule
    \end{tabular}
    \vspace{-0.3cm}
\end{table}

\begin{table}[t]
    \centering
    \caption{Ablation on the Mini-Batch Size $B$.}
    \label{tab:ablation_on_batch_size}
    \setlength{\tabcolsep}{2.0mm}
    \vspace{-0.2cm}
    \begin{tabular}{c|cc|cc|cc|cc}
        \toprule
        \multirow{2}{*}{$B$} & \multicolumn{2}{c|}{Pitts30k} & \multicolumn{2}{c|}{MSLS-val} & \multicolumn{2}{c|}{Nordland} & \multicolumn{2}{c}{Average} \\
        \cline{2-9}
        & R@1 & R@5 & R@1 & R@5 & R@1 & R@5 & R@1 & R@5 \\
        \hline
        120 & 92.5 & 96.6 & 92.4 & \textbf{96.4} & 87.2 & 94.9 & 90.7 & 96.0 \\
        160 & 92.7 & \textbf{96.8} & 92.4 & 96.1 & \textbf{88.3} & \textbf{95.3} & 91.1 & \textbf{96.1} \\
        \rowcolor{lightblue} 200 & \textbf{92.9} & \textbf{96.8} & \textbf{92.6} & \textbf{96.4} & \textbf{88.3} & 95.2 & \textbf{91.3} & \textbf{96.1} \\
        240 & 92.8 & 96.6 & 92.0 & 96.1 & 87.0 & 94.4 & 90.6 & 95.7 \\
        \bottomrule 
    \end{tabular}
    \vspace{-0.3cm}
\end{table}

\begin{table}[!ht]
    \centering
    \caption{Ablation on the Number of Nearest Neighbors $k$.}
    \setlength{\tabcolsep}{2.0mm}
    \label{tab:ablation_on_neighbors}
    \vspace{-0.2cm}
    \begin{tabular}{c|cc|cc|cc|cc}
        \toprule
        \multirow{2}{*}{$k$} & \multicolumn{2}{c|}{Pitts30k} & \multicolumn{2}{c|}{MSLS-val} & \multicolumn{2}{c|}{Nordland} & \multicolumn{2}{c}{Average} \\
        \cline{2-9}
        & R@1 & R@5 & R@1 & R@5 & R@1 & R@5 & R@1 & R@5 \\
        \hline
        1 & \textbf{92.9} & \textbf{96.8} & 92.3 & 95.9 & 85.7 & 93.9 & 90.3 & 95.5 \\
        \rowcolor{lightblue} 3 & \textbf{92.9} & \textbf{96.8} & \textbf{92.6} & \textbf{96.4} & 88.3 & 95.2 & \textbf{91.3} & \textbf{96.1} \\
        5 & 92.6 & 96.6 & 92.0 & 96.2 & 86.2 & 94.2 & 90.3 & 95.7 \\
        7 & 92.7 & 96.7 & \textbf{92.6} & 96.3 & 88.3 & 95.4 & 91.2 & \textbf{96.1} \\
        10 & 92.6 & 96.6 & 92.3 & 96.2 & \textbf{88.4} & \textbf{95.6} & 91.1 & \textbf{96.1} \\
        \bottomrule 
    \end{tabular}
    \vspace{-0.3cm}
\end{table}

\subsection{Additional Hyperparameter Analysis}
The proposed method mainly involves three hyperparameters: the weighting coefficient $\alpha$, the mini-batch size $B$ for scalable selection, and the number of nearest neighbors $k$ for estimating IPS. In the main paper, we set $\alpha=0.2$, $B=200$, and $k=3$ under the 30\% pruning ratio. Here, we provide additional ablation studies to examine the sensitivity of these hyperparameters. Unless otherwise specified, pruning is performed on GSV-Cities, and the final VPR model is DINOv2-B with NetVLAD, consistent with the corresponding setting in the main paper. The pruning ratio is fixed at 30\%.

Tables \ref{tab:ablation_on_alpha}, \ref{tab:ablation_on_batch_size}, and \ref{tab:ablation_on_neighbors} report the results. Overall, the proposed method remains stable across a range of hyperparameter settings. For $\alpha$, the best average performance is obtained at $\alpha=0.2$, while nearby values such as $\alpha=0.1$ and $\alpha=0.3$ also achieve competitive results. This indicates that both IPD and IPS contribute to coreset selection, and a moderate balance between them is generally effective. For the mini-batch size $B$, the performance remains close when $B$ varies from 120 to 240, showing that the proposed mini-batch selection strategy is not highly sensitive to the exact batch size. For the number of nearest neighbors $k$, the performance is not strictly monotonic with respect to the neighborhood size. Since IPS is estimated within each selection mini-batch, different values of $k$ may affect the locality of inter-place similarity estimation. In this 30\% pruning setting, $k=3$ achieves the best average performance, while several other settings remain competitive. In practice, we empirically choose suitable $k$ values for different pruning ratios instead of fixing a universal value across all settings. These results suggest that the proposed framework does not rely on delicate hyperparameter tuning, which is important for large-scale VPR training where exhaustive hyperparameter search would be costly.

\begin{table}[t]
    \centering
    \caption{Comparison of Different Dataset Pruning Methods with Various Backbones Under the 50\% Pruning Ratio. NetVLAD Is Used as the Aggregation Module.}
    \label{tab:cross_backbone_generalization}
    \setlength{\tabcolsep}{1.5mm}
    \vspace{-0.2cm}
    \begin{tabular}{c|c|cc|cc|cc}
        \toprule
        \multirow{2}{*}{Backbone} & \multirow{2}{*}{Method} & \multicolumn{2}{c|}{Pitts30k} & \multicolumn{2}{c|}{MSLS-val} & \multicolumn{2}{c}{Nordland} \\ 
        \cline{3-8}
        & & R@1 & R@5 & R@1 & R@5 & R@1 & R@5 \\
        \hline
        \rowcolor{gray!20} \cellcolor{white}
        \multirow{9}{*}{VGG-16} & full-dataset & 86.5 & 93.4 & 79.9 & 87.6 & 35.0 & 51.7 \\
        & Random & 85.8 & 92.4 & 76.9 & 85.0 & 28.0 & 41.4 \\
        & Herding & 85.7 & 92.3 & 75.5 & 84.7 & 28.0 & 41.4\\
        & K-Center & 85.8 & 92.5 & 77.8 & 85.5 & 28.9 & 43.2 \\
        & Moderate & \textbf{86.5} & 93.0 & 77.0 & 85.0 & 27.1 & 40.4 \\
        & CCS & 85.2 & 92.7 & 77.4 & 85.1 & 27.8 & 41.3 \\
        & FDMat & 85.4 & 92.5 & 77.6 & 85.7 & 27.9 & 41.7 \\
        & D2Pruning & 85.7 & 93.1 & \textbf{80.0} & \textbf{87.3} & 31.9 & 47.8 \\
        \rowcolor{lightblue} \cellcolor{white} & Ours & \textbf{86.5} & \textbf{93.3} & \textbf{80.0} & 87.2 & \textbf{33.2} & \textbf{49.1} \\
        \hline
        \hline
        \rowcolor{gray!20} \cellcolor{white}
        \multirow{9}{*}{ResNet-50} & full-dataset & 88.6 & 94.7 & 81.7 & 88.9 & 64.9 & 78.3 \\
        & Random & 87.6 & 94.4 & 79.3 & 85.7 & 52.5 & 68.4 \\
        & Herding & 86.9 & 93.8 & 78.8 & 85.8 & 51.5 & 67.6\\
        & K-Center & 86.8 & 93.8 & 79.9 & 86.5 & 51.6 & 66.8 \\
        & Moderate & 87.1 & 94.0 & 79.2 & 86.6 & 47.8 & 62.9 \\
        & CCS & 87.3 & 94.4 & 80.9 & 87.2 & 48.1 & 63.4 \\
        & FDMat & 87.2 & 93.7 & 80.5 & 86.5 & 50.6 & 66.4 \\
        & D2Pruning & 88.4 & 94.3 & 80.9 & 87.8 & 59.1 & 73.6 \\ 
        \rowcolor{lightblue} \cellcolor{white} & Ours & \textbf{88.5} & \textbf{94.5} & \textbf{81.4} & \textbf{88.2} & \textbf{63.3} & \textbf{77.1} \\
        \hline
        \hline
        \rowcolor{gray!20} \cellcolor{white}
        \multirow{9}{*}{ViT-B} & full-dataset & 90.6 & 95.6 & 85.7 & 92.4 & 55.1 & 70.4 \\ 
        & Random & 88.3 & 95.1 & 80.9 & 89.9 & 32.8 & 47.7 \\
        & Herding & 89.1 & 94.8 & 80.1 & 88.6 & 34.9 & 51.2\\
        & K-Center & 88.9 & 94.9 & 81.5 & 89.5 & 32.9 & 47.6 \\
        & Moderate & 89.2 & 94.9 & 80.7 & 89.1 & 32.0 & 47.4 \\
        & CCS & 89.5 & 94.8 & 82.2 & 89.5 & 38.3 & 52.8 \\
        & FDMat & 88.7 & 94.4 & 80.1 & 89.7 & 34.1 & 49.3 \\
        & D2Pruning & 90.2 & \textbf{95.7} & 84.6 & 91.2 & 51.5 & 66.2 \\ 
        \rowcolor{lightblue} \cellcolor{white} & Ours & \textbf{90.7} & 95.4 & \textbf{85.5} & \textbf{92.3} & \textbf{53.3} & \textbf{68.7} \\
        \bottomrule
    \end{tabular}
    \vspace{-0.3cm}
\end{table}

\subsection{Generalization Analysis Across Different Backbones}
In the main paper, we mainly evaluate the selected coresets using the DINOv2-B backbone. However, an effective dataset pruning method should not be overly tied to a specific network architecture. To further examine the generalization ability of the selected coresets, we conduct additional experiments with different backbones while keeping NetVLAD as the aggregation module.

\textbf{Implementation Details.} We evaluate three representative backbones, including VGG-16, ResNet-50, and ViT-B. All backbones are initialized with ImageNet-pretrained weights, and we follow the training protocols used in the Deep Visual Geo-localization Benchmark \cite{vprbenchmark2}. For VGG-16, only the final layer is trainable, while all preceding layers are frozen, and the initial learning rate is set to 0.0002. For ResNet-50, we crop the network at the second-to-last residual block to retain higher-resolution local features. Similar to VGG-16, only the final layer of the cropped network is trained, with an initial learning rate of 0.0002. For ViT-B, we remove the last two transformer blocks and keep all remaining blocks trainable, with the learning rate set to 0.0003. The same pruning protocol is used for all compared methods, and the final VPR models are trained on the corresponding selected coresets.

\textbf{Results and Analysis.} The results are reported in Table \ref{tab:cross_backbone_generalization}. Overall, our method achieves the best or comparable performance across different backbone architectures. Compared with random selection, our method consistently brings clear improvements. For example, with ViT-B, our method improves R@1 by 2.4, 4.6, and 20.5 percentage points on Pitts30k, MSLS-val, and Nordland, respectively. Compared with other dataset pruning baselines, our method also achieves the best average performance under all three backbones. These results indicate that the proposed place-wise scoring strategy can select informative coresets that are not restricted to a specific final VPR backbone. It is also worth noting that, compared with full-dataset training, our coreset achieves very similar performance on Pitts30k and MSLS-val, while showing a larger performance gap on Nordland. One possible reason is that Nordland has a larger domain gap from the predominantly urban training data. Backbones such as VGG-16, ResNet-50, and vanilla ViT-B generally have weaker transferable representations than recent large-scale pre-trained visual foundation models (e.g., DINOv2), and may therefore rely more heavily on the diversity and coverage of the full training set for cross-domain generalization. Nevertheless, our method still clearly outperforms other pruning baselines on Nordland, demonstrating that the selected coreset preserves more useful training places under different backbone architectures.

\subsection{Details of Benchmark Datasets}
\textbf{Pitts30k} \cite{pitts} is constructed from GPS-tagged Google Street View panoramas and provides 24 viewpoint samples per location in urban environments. The dataset features substantial viewpoint changes, moderate appearance variations, and limited dynamic content. It is a curated subset of the larger Pitts250k benchmark, while being more challenging for most approaches. In our experiments, we primarily evaluate on the Pitts30k-test, which is one of the most widely used datasets in the VPR task.

\textbf{MSLS} \cite{msls} is a comprehensive benchmark for VPR, containing over 1.6 million street-level images collected in 30 cities worldwide, covering urban, residential, and natural settings. Each image is accompanied by GPS metadata and heading information. The dataset captures a broad spectrum of real-world variations, including differences in lighting, weather, seasons, viewpoints, and the presence of moving objects. MSLS is organized into three partitions: a train split, a public validation set (MSLS-val), and a hidden test set (MSLS-challenge). Following prior practice \cite{salad,cricavpr,selavpr++}, our evaluation is conducted on both MSLS-val and MSLS-challenge to ensure a thorough and comparable assessment.

\textbf{Nordland} \cite{nordland_original,nordland} consists of video sequences recorded from a front-facing camera mounted on a train, covering the same route across four different seasons. The dataset contains substantial appearance changes due to seasonal and illumination shifts, while the viewpoint remains constant. It mainly features natural and suburban landscapes, and frame-level alignment is available as ground-truth correspondence. In line with prior work \cite{vprbenchmark2}, we sample frames at 1 FPS and adopt the winter images as the query set, with the summer images serving as the reference (database). 

Notably, all these datasets are organized according to the Deep Visual Geo-localization Benchmark \cite{vprbenchmark2} and then used for testing different models.

\subsection{Details of Compared Dataset Pruning Methods}
For a fair comparison, all compared methods are adapted to the place-wise VPR setting. Specifically, selection is performed on place-level descriptors rather than individual image descriptors. For methods that require class-wise statistics or difficulty scores, we construct the required statistics from place-level descriptors and use IPD as a VPR-compatible difficulty measure when necessary.

\textbf{Random.} Random selection uniformly samples places from the training set without using feature distributions or structural information. It has negligible selection cost but may fail to preserve informative or representative places.

\textbf{Herding} \cite{herding}. Herding aims to construct a coreset whose mean feature representation approximates that of the full dataset. It greedily selects samples to reduce the discrepancy between the coreset mean and the full-dataset mean in feature space. Although simple and deterministic, its iterative selection process can be costly on large-scale datasets.

\textbf{K-Center} \cite{k-center}. K-Center selects samples to maximize coverage in the feature space. It greedily chooses points such that each unselected sample is close to at least one selected point under a given distance metric. This encourages diversity and broad data coverage, but the repeated distance computation introduces considerable overhead when the dataset is large.

\textbf{Moderate} \cite{moderate_ds}.
Moderate selects samples whose feature distances to class centers are close to the median distance, assuming that such samples better represent the typical structure of each class. Since VPR does not provide conventional classification labels, we adapt Moderate by constructing center-based statistics from place-level descriptors.

\textbf{CCS} \cite{ccs}. Coverage-centric Coreset Selection first ranks samples according to their difficulty scores and then performs stratified sampling over different difficulty ranges. This strategy aims to maintain coverage across samples with different difficulty levels while avoiding over-selection of outliers. In our VPR adaptation, IPD is used as the difficulty score for place-level selection.

\textbf{FDMat} \cite{fdmat}. FDMat selects samples by matching feature distributions between the coreset and the full dataset. It estimates feature distribution statistics and formulates coreset selection as an optimal transport problem, which is solved approximately with the Sinkhorn algorithm. Similar to Moderate, FDMat relies on center-based statistics, and we adapt it to VPR using place-level descriptors and constructed centers.

\textbf{D2Pruning} \cite{d2pruning}. D2Pruning is a hybrid method that considers both sample difficulty and diversity. It builds a graph over the dataset and performs message passing for coreset selection. This graph-based procedure can be computationally expensive on large-scale datasets. Since common difficulty scores such as Forgetting \cite{forgetting} and EL2N \cite{el2n} are not naturally available in VPR, we use IPD as the difficulty measure when adapting D2Pruning to place-wise VPR pruning.

\subsection{Limitations and Future Directions}
Although the proposed VPR-oriented dataset pruning framework shows strong performance and scalability, there remain several practical considerations. Similar to many existing dataset pruning methods \cite{svp,moderate_ds,herding,k-center,fdmat,d2pruning}, our method relies on a proxy model to extract descriptors before coreset selection. This descriptor extraction step introduces an additional one-time cost. However, this cost is shared by the pruning pipeline and can be amortized once the selected coreset is reused for different VPR models or training settings. Moreover, as shown in our efficiency analysis, the subsequent selection step of our method is highly efficient and does not become the computational bottleneck.

Another practical issue lies in descriptor storage. When the training dataset is very large and the descriptor dimensionality is high, storing all pre-extracted descriptors may require non-negligible memory. This issue can be alleviated by using compact representations, lower-dimensional descriptors, or streaming/mini-batch computation strategies. Future work may further explore descriptor-efficient or descriptor-free pruning mechanisms to reduce the dependence on pre-extracted features. In addition, incorporating factors such as scene rarity and diversity may help construct more reliable coresets under extremely aggressive pruning ratios. We believe these directions can further improve the practicality and scalability of dataset pruning for large-scale VPR training.

\end{document}